\documentclass[11pt]{article}
\usepackage[english]{babel}
\usepackage[utf8x]{inputenc}
\usepackage{fullpage}
\usepackage{url}
\usepackage{color}
\usepackage{nk}
\usepackage{pdfpages}
\usepackage{amsmath}
\usepackage{enumitem}
\usepackage[colorlinks=true,
linkcolor=blue,
urlcolor=blue,
citecolor=blue,
plainpages=false]{hyperref}
\usepackage{wrapfig}
\usepackage{natbib}
\numberwithin{equation}{section}
\numberwithin{theorem}{section}
 
\usepackage[left]{lineno}
\usepackage{blindtext}
\usepackage{longtable}
\usepackage{multirow}

\newcommand*\patchAmsMathEnvironmentForLineno[1]{%
	\expandafter\let\csname old#1\expandafter\endcsname\csname #1\endcsname
	\expandafter\let\csname oldend#1\expandafter\endcsname\csname end#1\endcsname
	\renewenvironment{#1}%
	{\linenomath\csname old#1\endcsname}%
	{\csname oldend#1\endcsname\endlinenomath}}%
\newcommand*\patchBothAmsMathEnvironmentsForLineno[1]{%
	\patchAmsMathEnvironmentForLineno{#1}%
	\patchAmsMathEnvironmentForLineno{#1*}}%
\AtBeginDocument{%
	\patchBothAmsMathEnvironmentsForLineno{equation}%
	\patchBothAmsMathEnvironmentsForLineno{align}%
	\patchBothAmsMathEnvironmentsForLineno{flalign}%
	\patchBothAmsMathEnvironmentsForLineno{alignat}%
	\patchBothAmsMathEnvironmentsForLineno{gather}%
	\patchBothAmsMathEnvironmentsForLineno{multline}%
}

\usepackage{epstopdf}
\usepackage{tgpagella}
\DeclareMathAlphabet{\mathsf}{OT1}{cmss}{m}{n}
\SetMathAlphabet{\mathsf}{bold}{OT1}{cmss}{bx}{n}

\makeatletter

\newcommand{\Rmnum}[1]{\expandafter\@slowromancap\romannumeral #1@}

\makeatother

\makeatletter
\newcommand*{\rom}[1]{\expandafter\@slowromancap\romannumeral #1@}
\makeatother

\let\hat\widehat
\let\tilde\widetilde

\def\Sbar{{\bar S}}

\def\betah{{\hat \beta}}

\def\xih{{\hat \xi}}

\def\stilde{{\tilde s}}

\def\rhot{{\tilde \rho}}
\def\betat{{\tilde \beta}}

\newcommand{\op}{\mathrm{op}}
\newcommand{\var}{\mathrm{Var}}

\newcommand{\diam}{\rm{diam}}
\def\##1\#{\begin{align}#1\end{align}}
\def\$#1\${\begin{align*}#1\end{align*}}

\title{Detecting Nonlinear Causality in Multivariate Time Series with Sparse Additive Models}

\author{Yingxiang Yang, Adams Wei Yu, Zhaoran Wang, Tuo Zhao}

\begin{document}

\maketitle

\begin{abstract}
	We propose a nonparametric method for detecting nonlinear causal relationship within a set of multidimensional discrete time series, by using sparse additive models (SpAMs). We show that, when the input to the SpAM is a $\beta$-mixing time series, the model can be fitted by first approximating each unknown function with a linear combination of a set of B-spline bases, and then solving a group-lasso-type optimization problem with nonconvex regularization. Theoretically, we characterize the oracle statistical properties of the proposed sparse estimator in function estimation and model selection. Numerically, we propose an efficient pathwise iterative shrinkage thresholding algorithm (PISTA), which tames the nonconvexity and guarantees linear convergence towards the desired sparse estimator with high probability.
\end{abstract}

\section{Introduction}

Detecting causal relationships within a set of coupled stochastic processes has been an interesting and important problem that rises broadly across the fields of modern science and engineering, including financial industries \citep{atanasov2016shock}, computer science \citep{rodriguez2011uncovering}, epidemiology \citep{robins2000marginal}, neural science \citep{quinn2011estimating}, and climatology \citep{runge2015identifying}. A typical problem involving causal discovery often involves a vast network containing many agents, each generating a discrete time series. As the backbone that captures the evolutionary dynamics, the underlying causal relationships, which are often depicted by a causal graph \citep{quinn2015directed}, can provide guidance for estimating the values of the time series in near future. 

In this paper, we focus on learning Granger causality from a set of discrete time series when (i) the underlying causal graph is sparse, and (ii) when the underlying causal relationships are potentially nonlinear. We accomplish this goal by fitting the observed time series into a nonparametric sparse additive model (SpAM) of the form
\#\label{eq::cam}
X_i[t+1]=\sum_{j=1}^{p}f_{ij}(X_j[t])+\epsilon_i[t+1],
\#
where $X_i[t]$ is the $i$-th dimension of the $t$-th observation, and $\epsilon_i[t]$ is the additive noise. The sparsity refers to the fact that for each $i$ only a few $f_{ij}$'s are nonzero. We define $\calP_i = \{j:f_{ij}\not\equiv 0\}$.

\subsection{Motivations}

We motivate this work from several aspects, including why we focus on the notion of Granger causality, nonlinear causal relationships, and why we model this causal relationship using SpAMs.

{\bfseries\noindent Why Granger causality?} Granger causality \citep{granger1969investigating} is the most widely used notion of causality for time series observations. It captures the information flow between different dimensions \citep{kamitake1984time,rissanen1987measures}. In \eqref{eq::cam}, the existence of Granger causality can be directly interpreted in the sense of model selection: assuming that the density of $X_j[t]$ on the support of $f_{ij}$ is positive, then the conditional directed information from $\{X_j[t]\}_{t=1}^{n}$ to $\{X_i[t]\}_{t=1}^{n}$ given all other processes nonzero if and only if $f_{ij}\not\equiv 0$. 

{\bfseries\noindent Why nonlinear causality with SpAMs?} In many applications, nonlinear causal relationships rise naturally: in financial markets, it is well known that nonlinear causal relationship exists between the return and volume \citep{hiemstra1994testing}. In climatology, the causal relationships between adjacent as well as far apart geological regions have been shown to follow a nonlinear relationship \citep{runge2017detecting}. In those applications, merely considering models that only encode linear causality may result in estimation errors caused by model mismatches. More importantly, the limitation of linear models restricts our understanding of nonlinear dynamics within the evolution of the time series dataset. Therefore, studies on revealing nonlinear causal relationship behind a time series dataset is of great importance.

Most of existing works consider linear causal relationships within a generative model, such as the vector autoregressive models (VARs): $X[t+1]=\Ab X[t]+\epsilon[t+1]$, while the nonlinear causal relationships are often described by models with parametric forms \citep{fan2008nonlinear}. The fitting of nonparametric models, such as the nonparametric regression model $X[t+1]=f(X[t])+\epsilon[t+1]$, suffer from the curse of dimensionality \citep{fan2008nonlinear}. To the best of our knowledge, inferring nonlinear causality under nonparametric statistical models remain largely an open problem.

Among nonparametric statistical models, SpAMs is a less general but easier to fit model compared to the nonparametric regression model \citet{ravikumar2009sparse}. As a high dimensional extension to the nonparametric additive models \citep{hastie1990generalized}, SpAMs serve as an immediate generalization of the VARs that has direct implications on Granger causality. As such, we consider fitting SpAMs using time series data in this work.

\subsection{Related Works}

There exist several lines of research that are closely related to our work. In this section, we give a brief introduction to those, including the inference of Granger causality in large networks of stochastic processes, and the works related to fitting SpAMs.

{\bfseries\noindent Inferring Granger causality.} In a large network, two typical ways are used for detecting Granger causality: {\bf (i)} Granger causality is measured by information theoretic quantities, such as directed (or conditional directed) information \citep{quinn2015directed,rissanen1987measures,runge2017detecting}. While estimators with performance guarantee exist for the simple case \citep{jiao2013universal}, the intrinsic difficulty that lies beneath the computation of conditional directed information hinders the design of efficient algorithms that scale effectively in large networks. {\bf (ii)} The other way to detect Granger causality is under the assumption that the observations are generated from a statistical model. Several widely used models include the VAR \citep{han2013transition,etesami2014directed,materassi2010topological}, time series models with independent noise (TiMINo, \citet{peters2013causal}), and time series linear non-Gaussian acyclic models (TS-LiNGAM, \citet{hyvarinen2008causal}).

{\bfseries\noindent Fitting SpAMs.} SpAMs are direct extensions of nonparametric additive models \citep{hastie1990generalized} to the high dimensional regime. Common ways to fit SpAMs include: {\bf (i)} component selection and smoothing operator (COSSO, \citet{lin2006component}), which solves a group Lasso formulation in an reproducing kernel Hilbert space (RKHS), and {\bf (ii)} methods based on basis expansions, which expand each function $f_{ij}(\cdot)$ using a growing number of basis functions to reduce the problem into a group Lasso formulation \citet{HuangHW10}. An estimator based on kernel expansion is proposed in \citet{zhou2018non}. Compared to \citet{zhou2018non}, our work directly generalizes SpAMs and group Lasso with nonconvex regularization frameworks to fit time series data. The basis expansion method is superior compared to kernel expansion in the sense that sparse multiple kernel learning typically requires solving an optimization problem with $np$ parameters, with $p$ being the dimension and $n$ being the sample size. This is a significant bottleneck when analyzing large-scale, and high dimensional data. As we will explain later, our proposed method requires solving an optimization with $\calO(qp)$ parameters, with $q\ll n$.

\subsection{Our Contribution}

Our contribution in this paper is three-fold. Firstly, we generalize the fitting technique for SpAMs to the realm of time series data. By simple basis expansion, we construct a finite dimensional parametric approximation for each $f_{ij}$, and then estimate the parameters by solving a finite dimensional group-lasso-type optimization problem using nonconvex group regularization \citep{fan2001variable,yuan2006model,zhang2010nearly}. For the estimator we proposed, we provide statistical guarantees in terms of the oracle property: (I) For each $\{X_i[t]\}_{t=1}^{n}$, we correctly recover the set of $\{X_j[t]\}_{t=1}^{n}$'s that have causal effect on it; (II) The statistical rate of convergence in terms of function estimation is as small as directly performing low-dimensional estimation on the true support. Secondly, from a computational aspect, we propose an efficient pathwise iterative shrinkage thresholding algorithm (PISTA), which guarantees a linear convergence to the desired estimator with high probability. Lastly, we contribute to the field of causal inference by introducing a model for estimating nonlinear causality in high dimensional regime with both theoretical guarantees and fast numerical convergence. To the best of our knowledge, such models have only been evaluated empirically by \citet{peters2013causal} on low dimensional dataset.


\noindent\textbf{Notations}: Given a vector $v = (v_1, \ldots, v_{d})^\top\in \reals^{d}$, we denote the number of nonzero entries in $v$ as $\norm{v}_{0}$, and $v_{\setminus j} = (v_{1}, \ldots, v_{j-1}, v_{j+1}, \ldots, v_{d} )^\top \in \reals^{d-1}$ as the subvector of $v$ with the $j$-th entry removed. Let $\calA \subseteq \{1,...,d\}$ be an index set. We use $\bar{\Ac}$ to denote the complementary index set to $\calA$. Given a matrix ${\Ab}\in\reals^{d \times d}$, we use $\Lambda_{\max}({\Ab})$ and $\Lambda_{\min}(\Ab)$ to denote the largest and smallest eigenvalues of $\Ab$. Finally, to distinguish between SpAMs with \iid input, we refer to the model in \eqref{eq::cam} as SpAM with time series input (TS-SpAM).

\section{Fitting SpAMs for Time Series}

It is obvious that for a $p$-dimensional $X[t]$, fitting $p^2$ unknown functions $f_{ij}$'s can be performed in parallel by $p$ systems where each system fits \eqref{eq::cam} for a particular $i\in\{1,\ldots,p\}$. Therefore, for notational simplicity, we fix the subscript $i$ and denote $f_{ij}$ by $f_j$ in the rest of the paper. We also denote $\bf = [f_1,\ldots,f_p]\in\calF^p$.
Consider model \eqref{eq::cam} :

\begin{assumption}[General setup for TS-SpAM]\label{asm:gasm}
	~\\
	{\noindent (A1)} The cardinality of $\calP_i$ is uniformly upper bounded by $C_\calP<\infty$.
	
	{\noindent (A2)} For each $j\in[p]$, $f_{j}(x)$ is bounded on $\reals$ and belongs to the Sobolev space $W^{l,2}$.
	
	{\noindent (A3)} The noise $\epsilon[t]$ is bounded, and \iid across time, with $\Expect \epsilon[t]=0$ and $\Expect \epsilon^2[t]=\sigma^2$ for all $t\in[n]$.
	 
	{\noindent (A4)} Let $d=l+\alpha >0$, where $\alpha\in(0,1]$ and $l$ is a nonnegative integer such that $d>0.5$. The $l$-th derivative $f_{j}^{(l)}(\cdot)$ exists and satisfies a Lipschitz condition of order $\alpha$ for some constant $C<\infty$:
	$$|f_{j}^{(l)}(s)-f_{j}^{(l)}(t)| \le C|s-t|^\alpha, ~~\forall s,t\in[a,b],$$ where $[a,b]$ is the interval in which $X_j[t]$ takes value for any $t$. The boundedness of $X_j[t]$ follows from (A1) to (A3).
	
	{\noindent (A5)}  For $j\in\calP$, $\|f_{j}\|_2\geq c_f>0$ for some constant $c_f$.
	
	{\noindent (A6)} For each $i$, $X_i[t]$ has a distribution $Q_{i,t}(x)$, whose density $q_{i,t}(x)$ is positive over its support, and is bounded between $q_{\min}$ and $q_{\max}$ for all $i\in[p]$ and $t\in[n]$.
	
	{\noindent (A7)} The function $f_{j}(\cdot)$ and the distribution of $X_j[t]$ satisfies $\Expect f_{j}(X_j[t])=0$.
	
	{\noindent (A8)} For each $j$, we assume that $(X_j[t])_{t=1}^{n}$ is a $\beta$-mixing time series with $\beta\in(0,1)$ being a constant irrelevant of $p$. Furthermore, we assume that $\Expect \|X[1]\|<\infty$ where $\|\cdot\|$ denotes the standard Euclidean norm.
	
\end{assumption}

Most of the above assumptions are inherited from SpAM. The last assumption (A8) is common among works in time series. For simplicity, we restrict our attention to noise distributions with bounded support. The result can be easily extended to more general family of noise distributions such as those with subgaussian tails. The boundedness assumption on both $f_{j}(\cdot)$s and the noise allows us to work with a set of bounded discrete processes $\{X[t]\}_{t=1}^{n}$, which greatly simplifies our analysis.

We fit the TS-SpAM by solving a regularized optimization problem of the following form:
\begin{align}\label{eq:minf}
\min_{\bbf\in\calF^p}\frac{1}{n}\sum_{t=1}^{n}\left(Y[t]-\sum_{j=1}^{p}f_j(X_j[t])\right)^2+\Rc_{\lambda}(\bbf),
\end{align}
where $\calF$ consists of all the functions satisfying the constraints specified in Assumption \ref{asm:gasm}, and we denote the first term as the loss function $L_n(\bbf)$. 
Below, we introduce the detailed procedure for approximating each $f_i(\cdot)$ with splines, reducing the problem into a group-lasso-type optimization. The number of spline basis used is selected such that the approximation error balances the estimation error introduced when solving the group Lasso problem.
\subsection{Reduction to Group Lasso}\label{sec:redglasso}

Given $n$ samples, we construct a finite dimensional set $\calF_{jn}$ of splines to approximate $f_i(\cdot)$. Recall that $X_i[t]\in[a,b]$ by assumption. Given $n$ samples, partition the interval $[a,b]$ into $K+1$ sub-intervals, with $K=\calO(n^\nu)$ for $\nu\in(0,0.5)$. The end points of the partition are
$a=\zeta_0<\zeta_1<\cdots<\zeta_K<\zeta_{K+1}=b$, and are evenly located such that $\max_{1\le k\le K+1}|\zeta_k -\zeta_{k-1}| = \Oc(n^{-\nu})$. Let $\calF_{jn}$ be the set of polynomial splines that are $l-2$ times continuously differentiable on $(a,b)$, and are degree $l-1$ polynomials over each sub-interval $[\zeta_i,\zeta_{i+1}]$. The constant $l$ is defined in Assumption (A4). 

Under such settings, there exists a set of B-spline basis functions $\{\phi_k: 1\le k\le q \}$ of $\calF_{jn}$ where $q=K+l$. Such a set of bases can be constructed by recursively taking the divided difference from the Haar basis.\footnote{See Chapter 5's appendix of \cite{friedman2001elements} for a detailed description.} As additive models are prone to identifiability issues, \ie, $f_j$s are only identifiable up to constant differences, we center the splines to enforce the constraint that $\sum_{t=1}^{n}f_j(X_j[t])=0$:
\begin{align*}
\psi_{jk}(x):=\phi_k(x)-\frac{1}{n}\sum_{t=1}^{n}\phi_k(X_j[t]).
\end{align*}
Following the notations of \cite{HuangHW10}, we denote the space generated by the centered spline basis by $\calF_{jn}^0$.

At this point, we can approximate $f_{j}(x)$ in the model \eqref{eq::cam} by the functions in $\Fc_{jn}^0$:
\begin{align}\label{eq:basic}
f_{j}(x) \approx \sum_{k=1}^{q} \beta_{jk}\psi_{jk}(x), ~\forall j\in\calP.
\end{align}
Upon this approximation, the functional optimization problem \eqref{eq:minf} immediately turns into the following parametric estimation problem:
\begin{align}\label{eq:lsform}
\hat{\bbeta}\in\argmin_{\bbeta\in\reals^{pq}} &  {1\over 2n}\sum_{t=1}^{n}\Big(Y[t] - \sum_{j=1}^{p}\sum_{k=1}^{q}\beta_{jk}\psi_{jk}(X_{j}[t]) \Big)^2 + \Rc_\lambda(\bbeta),
\end{align}
where $\bbeta^\top=(\bbeta_{1}^\top,\bbeta_{2}^\top,\ldots,\bbeta_p^\top)$ is a $pq$-dimensional vector that concatenates $p$ column vectors, $\bbeta_i=(\beta_{i1},\ldots,\beta_{iq})^\top$, each of dimension $q$. More compactly,
\begin{align}\label{eq:beta_sam}
\hat{\bbeta}=\argmin_{\bbeta\in\reals^{pq}} {1\over 2n}\|y-Z\bbeta\|_2^2 +  \Rc_\lambda(\bbeta),
\end{align} 
where $y=(Y[1],Y[2],...,Y[n])^{\top}$ and $Z=(Z_1,...,Z_p)\in\reals^{n\times pq}$. Each component $Z_j$ is an $n\times q$ matrix, and is defined as $Z_j = (z_{1j}, z_{2j},...,z_{nj})^{\top}$ with $z_{ij}= (\psi_{j1}(X_{j}[i]),..., \psi_{jq}(X_{j}[i]))^\top$. More specifically, we have $Z\bbeta=\sum_{j=1}^{p}Z_j\beta_j$, where
\$
Z_j=
\begin{bmatrix}
	\psi_{j1}(X_j[1]) & \cdots & \psi_{jq}(X_j[1]) \\ \vdots & \ddots & \vdots \\ \psi_{j1}(X_j[n]) & \cdots & \psi_{jq}(X_j[n])
\end{bmatrix}.
\$

In \eqref{eq:lsform}, each group of coefficients $\{\beta_{jk}\}_{k=1}^{q}$ determines a function $f_{j}(\cdot)$. When $j\not\in\calP$, the recovered $\bbeta_{j}$ should be the zero vector. The imposed $\Rc_\lambda(\bbeta)$ is responsible for inducing group sparsity of the additive components. 

Finally, we characterize the approximation error of \eqref{eq:basic}, which parallels the result of approximation error when using the \iid samples in \cite{HuangHW10}.
\begin{lemma}[Approximation error with B-splines] \label{lm:approx}
	Under Assumption \ref{asm:gasm}, there exists $\tilde{f}_{j}\in\calF_{jn}^0$ such that
	%
	\begin{align*}
	\|f_{j}-\tilde{f}_{j}\|_2=\calO_P(q^{-d}+q^{1/2}n^{-1/2}).
	\end{align*}
	%
	
\end{lemma}
Given such a result, a natural way to choose $q$ is to balance the above two terms. Choosing $q=n^{1/(2d+1)}$ gives us $\|f_{j}-\tilde{f}_{j}\|_2=\calO_P(n^{-d/(2d+1)})$.

\subsection{Nonconvex Penalties}

We apply decomposable nonconvex penalty to \eqref{eq:beta_sam}. That is, \$\calR_\lambda(\bbeta)=\sum_{j=1}^{p}r_\lambda(\bbeta_j).\$
In particular, we focus on the nonconvex group MCP regularizer \citep{huang2012selective} of the form
\begin{align}\label{MCP-definition}
r_\lambda(\bbeta_j)&= \lambda\left(\norm{\bbeta_j}_2-\frac{\norm{\bbeta_j}_2^2}{2\lambda\gamma}\right)\cdot\mathds{1}(\norm{\bbeta_j}_2<\lambda\gamma) + \frac{\lambda^2\gamma}{2}\cdot\mathds{1}(\norm{\bbeta_j}_2\geq\lambda\gamma),
\end{align}
where $\gamma>0$ is the tuning parameter that determines the threshold on $\|\bbeta_j\|_2$ upon which $r'_\lambda(\bbeta_j)=0$. Note that $r_\lambda(\bbeta_j) = \lambda\|\bbeta_j\|_2 + w_\lambda(\bbeta_j)$ with $w_{\lambda}(\bbeta_j)$ being
\begin{align}\label{MCP-concave}
w_\lambda(\bbeta_j) &= \frac{\lambda^2\gamma-2\lambda\|\bbeta_j\|_2}{2}\cdot\mathds{1}(\|\bbeta_j\|_2\geq\lambda\gamma)-\frac{\|\bbeta_j\|_2^2}{2\gamma}\cdot\mathds{1}(\|\bbeta_j\|_2<\lambda\gamma).
\end{align}
Hence, the overall regularization term takes the form
$$\Rc_{\lambda}(\bbeta) = \sum_{j=1}^{p}[\lambda\|\bbeta_j\|_2 + w_\lambda(\|\bbeta_j\|_2)].$$

\subsection{Connection Between Time Series Observations and \iid Samples}

In this section, we make the observation that one can plug in the time series data into TS-SpAM as if they were \iid. The key to making such a connection resides in proving a restricted eigenvalue property for the block $Z_i$. This result also fuels our subsequent analysis of the oracle properties of our estimator with nonconvex regularization.

For strong mixing time series, the following Bernstein's inequality guarantees that the empirical average concentrates tightly around its expectation.

\begin{lemma}[Bernstein's inequality under strong mixing conditions \citep{merlevede2009bernstein}]\label{lm:Bernstein}
	Let $(X_j)_{j=1}^{\infty}$ be a sequence of centered random variables bounded by a constant $M$ and satisfying the strong mixing condition with parameter $\alpha_k\lesssim \exp(-\tilde{\mu}k)$. Then, there exists a $\tilde{\mu}$ dependent constant $C_{\tilde{\mu}}$, such that for all $n\geq 4$ and $x\geq 0$,
	\begin{small}
		\begin{align*}
		\Prob\lbrac \left|\sum_{i=1}^{n}X_i\right|\geq x\rbrac\leq 2\exp\lbrac-\frac{C_{\tilde{\mu}}x^2}{nM^2+Mx(\log n)(\log\log n)}\rbrac.
		\end{align*}
	\end{small}
	
\end{lemma}

The above lemma is similar to the Bernstein's inequality for \iid random variables. It turns out that this lemma will help us prove the restricted eigenvalue condition for the design matrix $Z$, which is the key to the theoretical results under the \iid scenario. We state this restricted eigenvalue condition for time series in the following lemma:

\begin{lemma}
	\label{lm:rip}
	Let $q=\calO(n^{\nu})$ where $\nu\in(0,1/2)$ and $n\geq 5$. For each $Z_i$, let $U_i=Z_i^{\top}Z_i/n$. Then with high probability, we have $$C_{\min}q^{-1}\leq\min_{i\in[p]}\rho_{\min}(U_i)\leq\max_{i\in[p]}\rho_{\max}(U_i)\leq C_{\max} q^{-1},$$ where $C_{\min}$ and $C_{\max}$ are constants dependent only on $\tilde{\mu}$ defined in Lemma \ref{lm:Bernstein}.
\end{lemma}



We briefly sketch the proof of Lemma \ref{lm:rip} below. A detailed proof can be found in the Appendix.

\begin{proof}[Proof Sketch]
	
	We relate the eigenvalues of $U_i$ to the maximum and minimum values of $v^\top Z_i^\top Z_i v$, where $\|v\|_{2}=1$ lies on the surface of a $(q-1)$-dimensional sphere $\mathbb{S}^{q-1}$.
	
	\emph{Upper bound.} The upper bound for $\rho_{\max}(U_i)$ can be obtained in three steps.
	
	(i) We first derive an exponential tail bound for $v^\top Z_i^\top Z_iv-\Expect v^\top Z_i^\top Z_iv$ for fixed $v\in\mathbb{S}^{q-1}$. This can be obtained by first noticing that
	\begin{align*}
	v^\top Z_i^\top Z_iv=\sum_{t=1}^{n}\Big(\sum_{j=1}^{q}\phi_j(X_i[t])v_j\Big)^2,
	\end{align*}
	and applying Lemma \ref{lm:Bernstein}. The Bernstein inequality requires that $\{(\sum_{j=1}^{q}\phi_j(X_i[t])v_j)^2\}_{t=1}^{\infty}$ be strong mixing. This is true since $(X_i[t])_{t=1}^{\infty}$ is strong mixing.
	
	(ii) Upper bounding $\Expect v^\top Z_i^\top Z_iv$ by $\calO(nq^{-1})$. This holds true by the property of the B-splines. Together with the first step, one has
	\begin{align*}
	\Prob\lbrac\frac{|S_{n}(v)|}{n}\geq \delta\rbrac&\leq 2\exp\lbrac-\frac{C_{\tilde{\mu}}n\delta^2}{M^2+M\delta\log n\log\log n}\rbrac.
	\end{align*}
	where $S_{n}(v)$ is the simplified notation of $v^\top Z_i^\top Z_iv-\Expect v^\top Z_i^\top Z_iv$. 
	
	(iii) We apply the standard $\epsilon$-net argument. Choose a proper value of $\epsilon$, and let $x=C_2q^{-1}$. When $q\asymp n^{-\nu}$ with $\nu\in(0,1/2)$, the probability that the supremum of $v'Z_i'Z_iv/(n-1)$ over $v$ converges to 0. Hence with high probability, $\|Z_i\|_{op}^2\leq C_{\max}nq^{-1}$.
	
	\emph{Lower bound.} To obtain the lower bound, one needs to obtain a lower bound for $\Expect [\|Z_iv\|^2]$. This can be done following the argument in Lemma 6.1 in \citet{zhou1998local}, where
	\begin{align*}
	\Expect[\|Z_iv\|^2]=\sum_{t=1}^{n-1}\int_{a}^{b}s_t^2(x)dQ_{i,t}(x)\geq nq_{\min}cq^{-1}\|v\|_2^2,
	\end{align*}
	where $c$ is some constant.
\end{proof}

Since the cardinality of $\calP$ is bounded by $C_{\calP}$, the following result holds true by triangle inequality and Lemma 3 of \cite{stone1985additive}:

\begin{lemma}[Restricted eigenvalues]\label{lm:rse}
	Let $q=\calO(n^{\nu})$, where $\nu\in(0,1/2)$, and $A\subseteq\{1,2,\ldots,p\}$ satisfying $|A|\leq C_\calP$. Define $U_A=Z_A^{\top}Z_A/n$. Then, with high probability, we have \$C_{\min}'q^{-1}&\leq\min_{A\subset\{1,\ldots,p\},|A|\leq\calP}\rho_{\min}(U_A)\quad\text{and} \quad C_{\max}' q^{-1}\geq\max_{A\subset\{1,\ldots,p\},|A|\leq\calP}\rho_{\max}(U_A),\$ where $C_{\min}'$ and $C_{\max}'$ are constants dependent only on $\tilde{\mu}$.
\end{lemma}


\section{Main Results: Oracle Properties} \label{sec:op}

We now introduce the oracle properties of the group lasso problem: when using a nonconvex regularization, the convergence rate of the estimation of $\bbeta$ will be as good as the convergence rate when the set $\calP$ is given by an oracle. As a matter of fact, the result presented below not only applies to the analysis of TS-SpAM, but also serves as a topic general enough to be of its own independent interest.

We first define the oracle solution of the group Lasso problem:
\begin{definition}[Oracle solution]
	For any given parent set $\calP$, let $\bbeta_{\calP}\in\reals^{|\calP|q}$ be the concatenation of $\bbeta_j$s with $j\in\calP$. Then, the oracle solution $\hat{\bbeta}^o$ is defined in the way such that $\hat{\bbeta}^o_j=\zero$ for $j\in\bar{\calP}$ and
	\begin{align}\label{eq:oracle}
	\hat{\bbeta}^o_{\calP}&\in \argmin_{\bbeta_\calP\in\reals^{|\calP|q}} {1\over 2n}\left\|y-\sum_{j\in\calP}Z_j\bbeta_j\right\|_2^2.
	\end{align}
\end{definition}
It is equivalent to saying that an oracle provides information on the support of $\hat{\bbeta}^o$, and therefore the regression is limited to estimating $\bbeta_j$ for $j\in\calP$, which is a low dimensional problem since $|\calP|$ is fixed and does not grow with $p$ or $n$.

For regularizers of the form defined in \eqref{MCP-definition}, such as group SCAD and group MCP \citep{huang2012selective}, the following regularity conditions on the concave function $w_\lambda(z): \reals\rightarrow\reals$, defined in \eqref{MCP-concave}, guarantee consistent estimation given \iid observations:
\begin{assumption}
	We require $w_{\lambda}(z)$ to satisfy the following regularities:
	\begin{itemize}
		\item[(R.a)] $w_\lambda$ is symmetric, i.e., $w_\lambda(z) = w_\lambda(-z)$.
		\item[(R.b)] $w_\lambda$ and $w_\lambda'$ pass through the origin, i.e. $w_\lambda(0)=w_\lambda'(0)=0$.
		\item[(R.c)] There is some $\gamma >0$, such that $\forall z\ge 0$, we have $w_\lambda'(z)\in [0,\lambda]$ if $z\le \lambda\gamma$, and $w_\lambda'(z)=-\lambda$ otherwise.
		\item[(R.d)] $w_\lambda'(z)$ is monotonically decreasing and Lipschitz continuous, i.e., for $z_1>z_2$, there are positive real numbers $\eta_-$ and $\eta_+$, such that $-\eta_-<(w_\lambda'(z_1) - w_\lambda'(z_2))(z_1-z_2)^{-1}<-\eta_+ <0$.
		\item[(R.e)] $w_\lambda'$ has bounded difference with respect to $\lambda$, i.e., $|w_{\lambda_1}'(z) -w_{\lambda_2}'(z)|\leq |\lambda_1-\lambda_2|$ for all $z$.
	\end{itemize}
\end{assumption}

Under the regularity assumptions on the penalty function and the general assumptions we made on TS-SpAM, the following oracle property holds for time series input.

\begin{theorem}\label{thm:recover}
	Given $q=\calO(n^{1/(2d+1)})$, {suppose that} there exists a universal constant $C$ such that
	\begin{align}\label{strong-signal}
	\min_{j\in S} \|\bbeta^*_j\|_2 \ge C\sigma \left(\sqrt{q\over n} + {\sqrt{\log p}\over n} \right).
	\end{align}
	If we set the regularization coefficient $\lambda_n = 8\sigma( \sqrt{1/n} + {\sqrt{\log p}/n})$, then for large enough $n$, we have
	\begin{align*}
	\Pb(\hat{\bbeta} =\hat{\bbeta}^o) \ge 1-p^{-1},
	\end{align*}
	while
	\$
	\norm{\hat{f}_j-f_j}_2^2=\calO_P(n^{-2d/(2d+1)}).
	\$
\end{theorem}

Theorem \ref{thm:recover} provides statistical guarantee of the probability for successfully performing support recovery as well as the $L_2$ error of function estimation. This result provides theoretical guarantee for inferring Granger causality in nonlinear autoregressive models.

\vspace{-0.1in}
\section{Computational Algorithm}\label{sec:pista}
\vspace{-0.05in}

Nonconvex optimization problems are computationally intractable in general, as has been extensively studied by the worst-case analysis in classical optimization theory. However, the nonconvex optimization  in our problem is not constructed adversarially, and thus one can naturally expect a better result for any computational algorithms when comparing to the worst-case performance. In this section, we develop an efficient pathwise iterative shrinkage thresholding algorithm (PISTA), which guarantees linear convergence to the desired sparse local solution with high probability. PISTA contains two loops: (i) the inner loop, which is an iterative shrinkage thresholding algorithm; (ii) the outer loop, which is the pathwise optimization scheme.

For notational convenience, we define the augmented loss function as $\tilde{L}_{\lambda}(\bbeta) = L(\bbeta)+ \Wc_{\lambda}(\bbeta)$, where $\Wc_{\lambda}$ is the summation of $w_{\lambda}$s defined in \eqref{MCP-concave}. With this notation, we rewrite the objective function as
\begin{align*}
F_\lambda(\bbeta) =\tilde{L}_{\lambda}(\bbeta) + \lambda\norm{\bbeta}_{1,2}.
\end{align*}

\subsection{Iterative Shrinkage Thresholding Algorithm}

We first derive the iterative shrinkage thresholding algorithm (ISTA), the inner loop of PISTA. Suppose that we have the solution $\bbeta^{(t)}$ at the $t$-th iteration. We take the following quadratic approximation of $F_{\lambda}(\bbeta)$ at $\bbeta = \bbeta^{(t)}$:
\begin{align*}
H_{\lambda,L_{t+1}}(\bbeta,\bbeta^{(t)}) &= \tilde{L}(\bbeta^{(t)}) + (\bbeta-\bbeta^{(t)})^\top\nabla\tilde{L}(\bbeta^{(t)})+ \frac{\eta_{t+1}}{2}\norm{\bbeta-\bbeta^{(t)}}_2^2 + \lambda\norm{\bbeta}_{1,2},
\end{align*}
where $\eta_{t+1}$ is the step size parameter for the $(t+1)$-th iteration. The ISTA algorithm then takes
\begin{align}\label{eq:proximal-gradient}
\bbeta^{(t+1)} &= \argmin_{\bbeta}~H_{\lambda,\eta_{t+1}}(\bbeta,\bbeta^{(t)}) =\argmin_{\bbeta}\frac{\eta_{t+1}}{2}\norm{\bbeta-\tilde{\bbeta}^{(t+1)}}_2^2 + \lambda\norm{\bbeta}_{1,2},
\end{align}
where $$\tilde{\bbeta}^{(t+1)} = \bbeta^{(t)} - \frac{1}{\eta_{t+1}}\nabla \tilde{L}(\bbeta^{(t)}).$$ By group soft thresholding, \eqref{eq:proximal-gradient} admits a closed form solution $\bbeta^{(t+1)}$, where $$\bbeta^{(t+1)}_j = \tilde{\bbeta}_j^{(t+1)}\cdot\max\left(1-\frac{\lambda}{\eta_{t+1}\norm{\tilde{\bbeta}_j^{(t+1)}}_2},~0\right).$$

Once again, for notational simplicity, we write $\bbeta^{(t+1)} = T_{\lambda,\eta_{t+1}}(\bbeta^{(t)})$. The step size $1/\eta_{t+1}$ in \eqref{eq:proximal-gradient} can be obtained by performing a backtracking line search. Particularly, we start with a small enough $\eta_0$, and then choose the minimum nonnegative integer $z$ such that $\eta_{t+1} = 2^{z}\eta_{t}$ satisfies $F_{\lambda}(\bbeta^{(t+1)}) \leq H_{\lambda,\eta_{t+1}}(\bbeta^{(t+1)},\bbeta^{(t)})$ for $t=1,2,\ldots$. We terminate ISTA when the following approximate KKT condition holds:
\begin{align*}
\kappa_\lambda(\bbeta^{(t+1)}) & = \min_{\xi\in\partial\norm{\bbeta^{(t+1)}}_{1,2}}\norm{\nabla \tilde{L}_{\lambda}(\bbeta^{(t+1)}) + \lambda\xi}_{\infty,2} \leq \varepsilon \lambda,
\end{align*}
where $\varepsilon$ is the target precision. 

\subsection{Pathwise Optimization Scheme}

We next derive the pathwise optimization scheme, the outer loop of PISTA. The pathwise optimization scheme is very natural for sparse optimization problems, since we always need to tune the regularization parameter $\lambda$ over a refined grid to achieve a good empirical performance. The pathwise optimization scheme takes advantage of the tuning procedure, and does not require any additional computational effort. Particularly, the pathwise optimization scheme considers a decreasing sequence of regularization parameters $\{\lambda_M\}_{M=0}^N$, where $\lambda_M = 0.95\lambda_{M-1} = 0.95^{\{M\}}\lambda_0$ for $M=1,...,N$. We can verify that when we choose $\lambda_0=\norm{\frac{1}{n}Z^\top y}_{\infty,2}$, $\bbeta=0$ is a local solution, i.e., $\hat{\bbeta}^{\{0\}} = 0$ and $\kappa_{\lambda_0}(0)=0$. Then for $M=1,2,....,N$, we adopt the ISTA algorithm with $\hat{\bbeta}^{\{M-1\}}$ as the initial value, and solve the nonconvex optimization problems with respect to $\lambda_M$. We summarize the PISTA algorithm in Algorithm 1.

\begin{algorithm}[!htb]
	\caption{The pathwise iterative shrinkage thresholding algorithm (PISTA) exploits the solution sparsity based on the pathwise optimization scheme. The large regularization parameters suppress the overselection of the nonzero block of coordinates, and yield large estimation bias. But the bias can be further reduced by the decreasing regularization sequence. Eventually, PISTA gradually recovers the relevant block of coordinates, reduces the estimation error of each output solution, and attains a sparse output solution with the desired statistical properties.}
	\begin{algorithmic}
		\STATE \textbf{Input:} $\{\lambda_{M}\}_{M=0}^N,~\varepsilon$ \hfill{\it \small Input Regularization Sequence}
		\STATE \textbf{Parameter:} $\varepsilon$
		\STATE \textbf{Initialize:} $M\leftarrow0,~\hat{\bbeta}^{\{0\}} \leftarrow 0$
		\STATE \textbf{For} M = 1, 2, ..., N\hfill{\it \small Pathwise Optimization Scheme}
		\STATE \hspace{0.2in} $\bbeta^{(0)} \leftarrow  \hat{\bbeta}^{\{M-1\}},~t \leftarrow  0$
		\STATE \hspace{0.2in}\textbf{Repeat:}
		\STATE \hspace{0.4in} $\bbeta^{(t+1)}\leftarrow T_{\lambda,\eta_{t+1}}(\bbeta^{(t)})$\hfill{\it \small Group Thresholding + Line Search}
		\STATE \hspace{0.4in} $t \leftarrow  t+1$
		\STATE \hspace{0.2in} \textbf{Until:} $\kappa_{\lambda}(\bbeta^{(t+1)}) \leq \varepsilon\lambda_M$ \hfill{\it \small Approximate KKT condition}
		\STATE \hspace{0.2in} $\hat{\bbeta}^{\{M\}} \leftarrow \bbeta^{(t+1)}$
		\STATE \textbf{End For}
		\STATE \textbf{Return:} $\{\hat{\bbeta}^{\{M\}}\}_{M=1}^N$ \hfill{\it \small Output Solution Sequence}
	\end{algorithmic}
\end{algorithm}


We establish the global convergence of PISTA in the next theorem.
\begin{theorem}\label{computational-theorem}
	Under the same assumptions as Theorem \ref{thm:recover}, and given $\lambda_N = 8\sigma \left( \sqrt{1/n} + {\sqrt{\log p}/n}\right)$, for large enough $n$, $M=1,...,N$, the following results hold with high probability:
	\begin{description}
		\item (1) Throughout all iterations, any iterate in our algorithm $\bbeta$ is sufficiently sparse, $\norm{\bbeta}_0\leq C_\calP/2$.
		\item (2) PISTA converges to a unique sparse local solution $\bar{\bbeta}^{\{M\}}$ for each $\lambda_M$ satisfying \begin{align*}
		\norm{\bar{\bbeta}^{\{M\}}}_0\leq C_\calP/2\quad\textrm{and}\quad\kappa_{\lambda_M}(\bar{\bbeta}^{\{M\}})=0.
		\end{align*}
		\item (3) For each $\lambda_M$, we have $\norm{\hat{\bbeta}^{\{M\}}-\bar{\bbeta}^{\{M\}}}_2^2=\calO\left(q C_\calP\varepsilon\right)$.
		\item (4) To compute the entire solution path, the number of ISTA iterations is at most $\calO(N\cdot\log (1/\varepsilon))$.
	\end{description}	
\end{theorem}

Theorem \ref{computational-theorem} is a result that extends the Theorem \ref{thm:recover} in \cite{wang2014optimal} to the group Lasso setting. In terms of the performance, Theorem \ref{computational-theorem} guarantees that PISTA attains global linear convergence to the sparse unique oracle solution for nonconvex structured causal inference problem with high probability. Note that $\bar{\bbeta}^{\{N\}}$ in Theorem \ref{computational-theorem} is exactly $\hat{\bbeta}$ in Theorem \ref{thm:recover}, since they correspond to the same value of $\lambda$.

\section{Numerical Results}

In this section, we verify the proposed algorithm on both synthetic and real data. For the synthetic data set, we verify the support recovery performance on a nonlinear autoregressive model with $f_{ij}$'s designed as polynomial splines. For the real data set, we infer the causal relationship from daily stock market price data of 452 companies listed at NYSE, and compare prediction error of PISTA and Lasso (realized using glmnet \citep{friedman2010regularization}). In particular, we demonstrate, from the real data, the superiority of PISTA over Lasso when the input data contains nonlinear causal relationship.

\subsection{Synthetic Data}

Consider the model \eqref{eq::cam}, where $p=300$, and $n=500$. \footnote{We set the number of spline bases for each dimension to be 3, so that the design matrix $Z$ has less number of samples than the number of predictors.} We focus on estimating $f_{ij}$'s for $i=1$, and we assume 10 out of the remaining 299 $f_{1j}$'s are non-zero, which we choose uniformly at random. For each $f_{1j}\not\equiv 0$, we set it to be of the form $f_{1j}(x) = a_jx+b_jx^2+c_jx^3$, where $[a_j,b_j,c_j]$ is a standardized vector that guarantees the mixing of $\{X_1[t]\}_{t=1}^{n}$. The noise $\epsilon_i[t]\sim\calU(-0.4,0.4)$. Finally, for $i\in\{2,\ldots,p\}$, we set $f_{ii}$ in the same fashion as the non-zero $f_{1j}$'s, and set $f_{ik}\equiv 0$ for $k\neq i$.

\begin{figure}
	\centering
	\begin{tabular}{cc}		
		\includegraphics[width=0.45\textwidth]{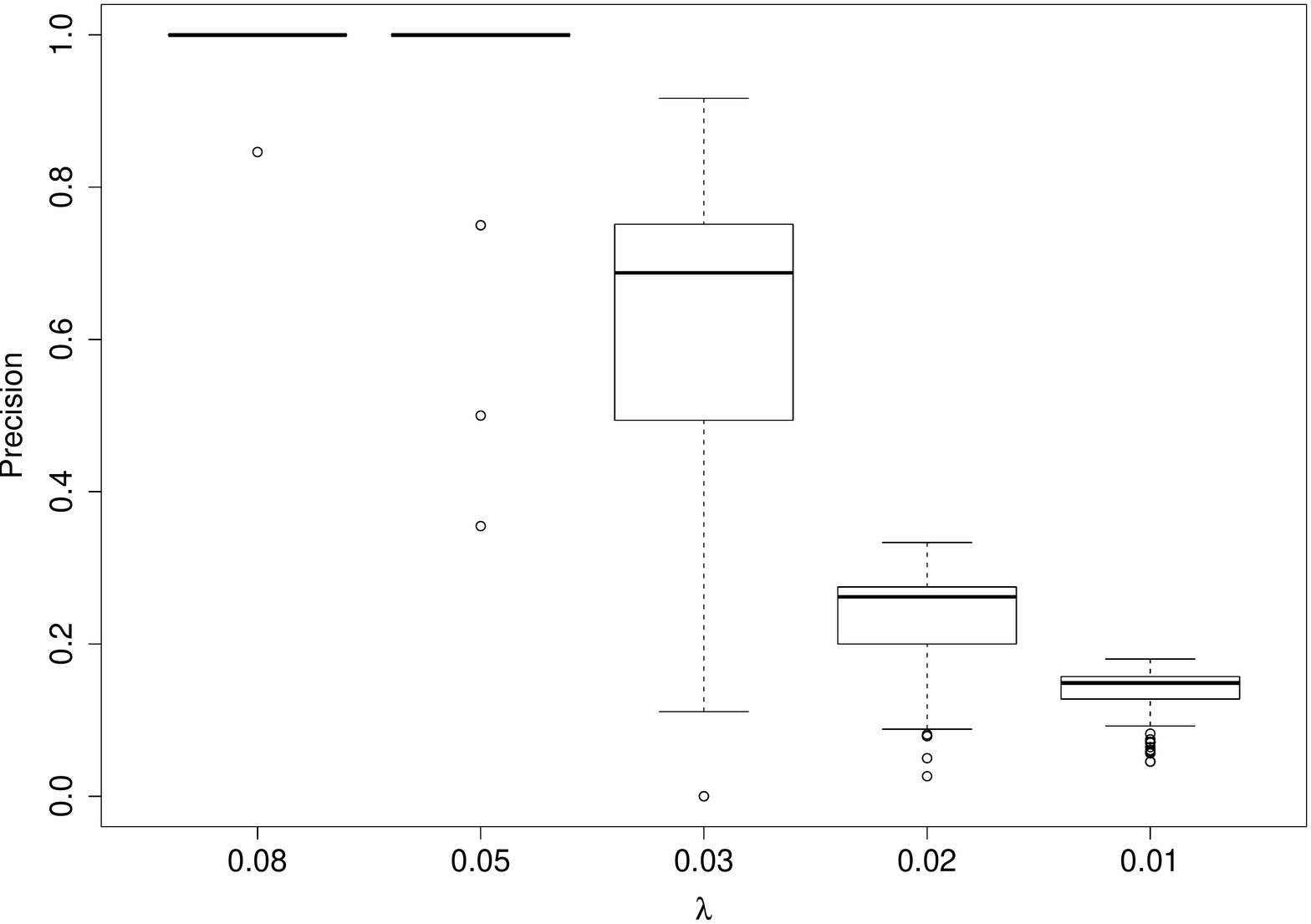}
		&\includegraphics[width=0.45\textwidth]{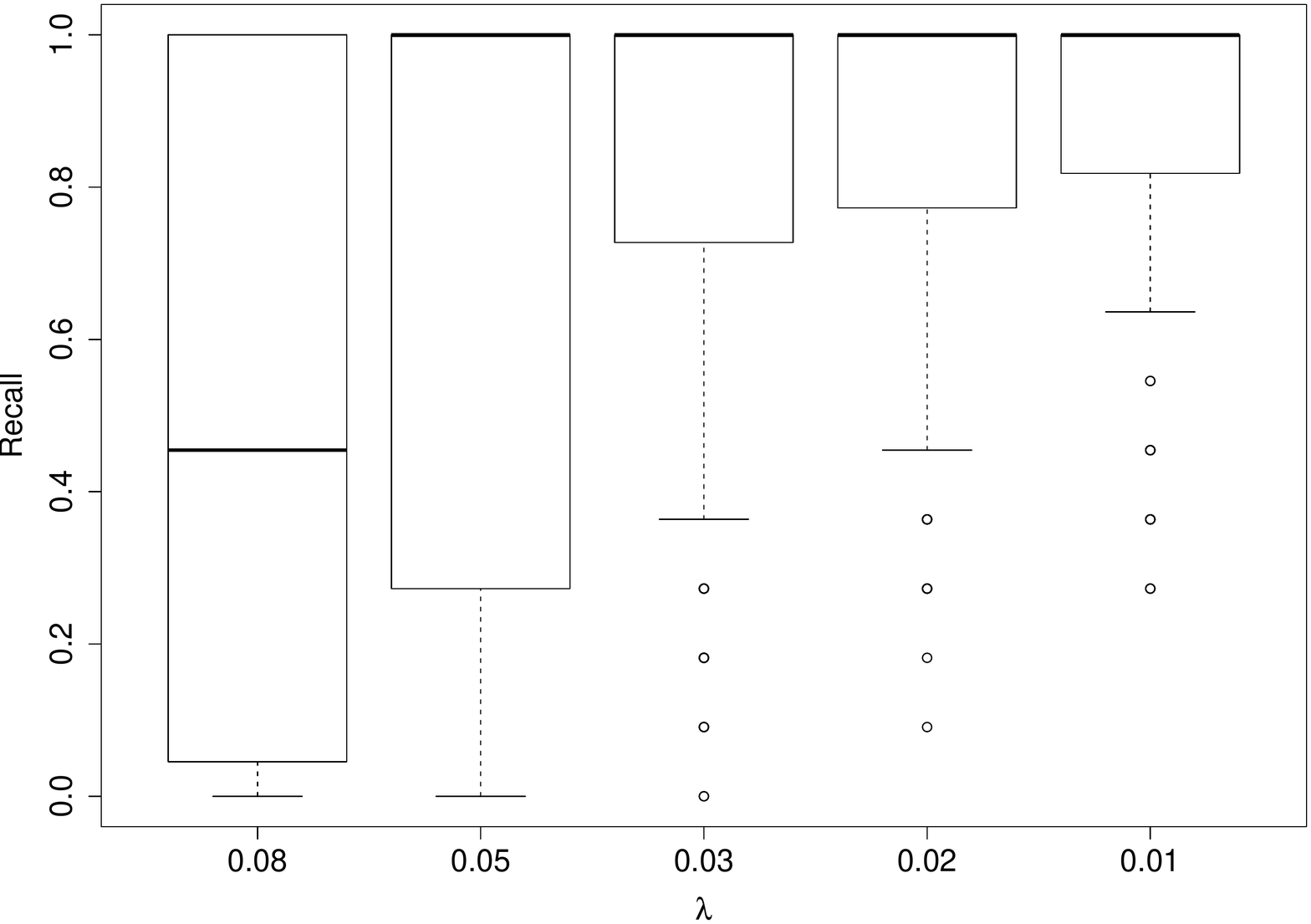}
	\end{tabular}
	\caption{Box plots of PISTA's performance on precision (left) and recall (right).}
	\label{fig::precrec}
\end{figure}

We randomly generate 100 training sets, and test PISTA's performance for different $\lambda$'s. \footnote{Here $\lambda$ refers to the maximum $\{\lambda_M\}_{M=1}^{N}$ specified as the input to PISTA.} We set the number of splines to be 3 and adopt the group MCP penalty with $\gamma=1$. \footnote{As shown in Appendix \ref{sec::moreplots}, increasing $\gamma$ does not significantly affect PISTA's performance in this case.} The computed precision and recall statistics are shown in Figure \ref{fig::precrec}. As $\lambda$ increases, the precision increases at the expense of lowering the recall. The balance of the two that maximizes the F1 score is achieved at roughly $\lambda\approx 0.045$ for PISTA, with an F1 score of $0.8$. The selected $\lambda$ generalizes from the training set to the test set so that optimal $\lambda$ can be selected via cross validation. In Appendix \ref{sec::moreplots}, we perform a side by side comparison between the precision and recall performance of PISTA and that of group Lasso. The result shows that the PISTA with group MCP penalty achieves much better performance in support recovery compared to group Lasso.

\subsection{Real Data: Inferring Causality From Stock Market Data}

We demonstrate and interpret the result of TS-SpAM when applied to a stock market dataset. The dataset we use is collected from \citet{zhao2012huge}, which includes 452 company's stock prices over a period of 1258 days between Jan 1st, 2003 to Jan 1st, 2008. In our experiment, we consider those stocks that belong to the ``information technology" sector, and we consider the first 100 samples to interpret the causal relationship.

We preprocess the data by considering the log return. The log return, $\{r_i[t]\}_{t=1}^{n}$, of a stock with a price history $\{X_i[t]\}_{t=1}^{n}$ is defined as
\#\label{eq::logreturn}
r_i[t] = \log \left(1+\frac{X_i[t]-X_i[t-1]}{X_i[t-1]}\right).
\#
For this dataset, the log return of each stock can be viewed as a stationary time series.

We then input the processed stock prices to PISTA with the number of B-spline basis set to be 3. We also input the processed stock prices to glmnet \citep{friedman2010regularization}, which we use as our benchmark. For both algorithms, we set the regularization parameter sequence to have a wide range such that the resulting estimated support could be sparse. For each stock, we acquire three companies that have the most significant impact on it. The complete estimated relationship can be found in Table \ref{tab::stock}, and we manually choose 24 dimensions to visualize in the following graphs.

The selected companies' labels are given in Table \ref{tab::tickers}. The majority of the companies we selected are related to the manufacturing of semiconductors or storages. These include NVDA, MU, AMD, INTC, QCOM, BRCM, SNDK, WDC, etc. A few companies are related to the manufacturing of computers and computer related softwares, including DELL, MSFT, HPQ, etc. The results of PISTA and glmnet have a lot in common. For example, SNDK, NVDA, QCOM, WDC are widely connected in the results of both algorithms. However, the result of PISTA also captures a few relationships that are not present in the glmnet result. For example, the result of PISTA shows that the price of MU causally affects the prices of NVDA, AMD, both of which are popular stocks in the semiconductor industry. Glmnet on the other hand, only captures the impact on INTC. \footnote{In our result, we keep the $\calP_i$ to be the same for glmnet and PISTA. Besides INTC, glmnet also recovers Molex and Paychex as the stocks significantly impacted by MU. While Molex is involved in manufacturing of electronics and fiber, Paychex is not directly related to semiconductor industry.}

\begin{figure}
	\centering
	\begin{tabular}{cc}		
		\includegraphics[width=0.5\linewidth]{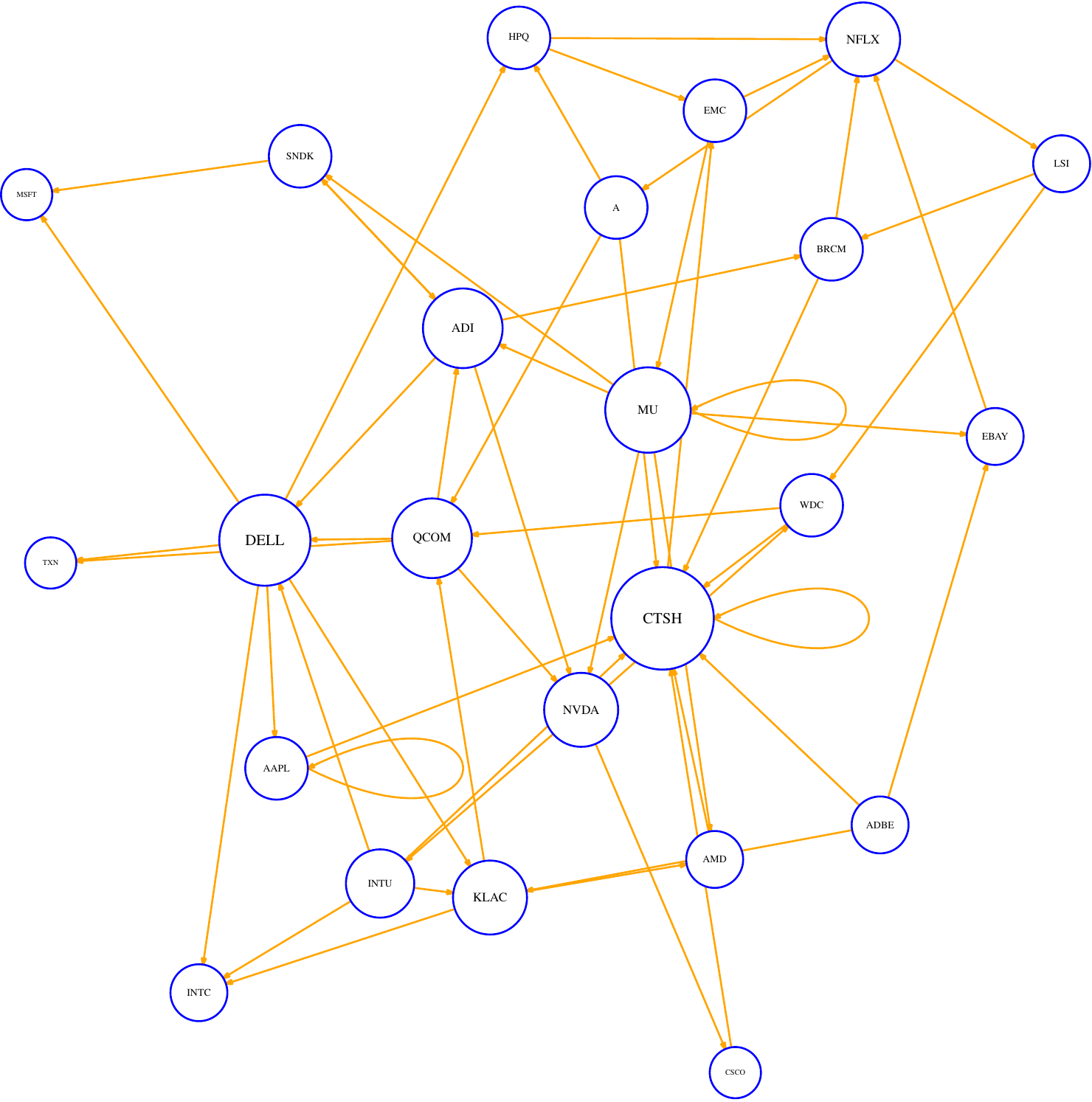}
		&\includegraphics[width=0.5\linewidth]{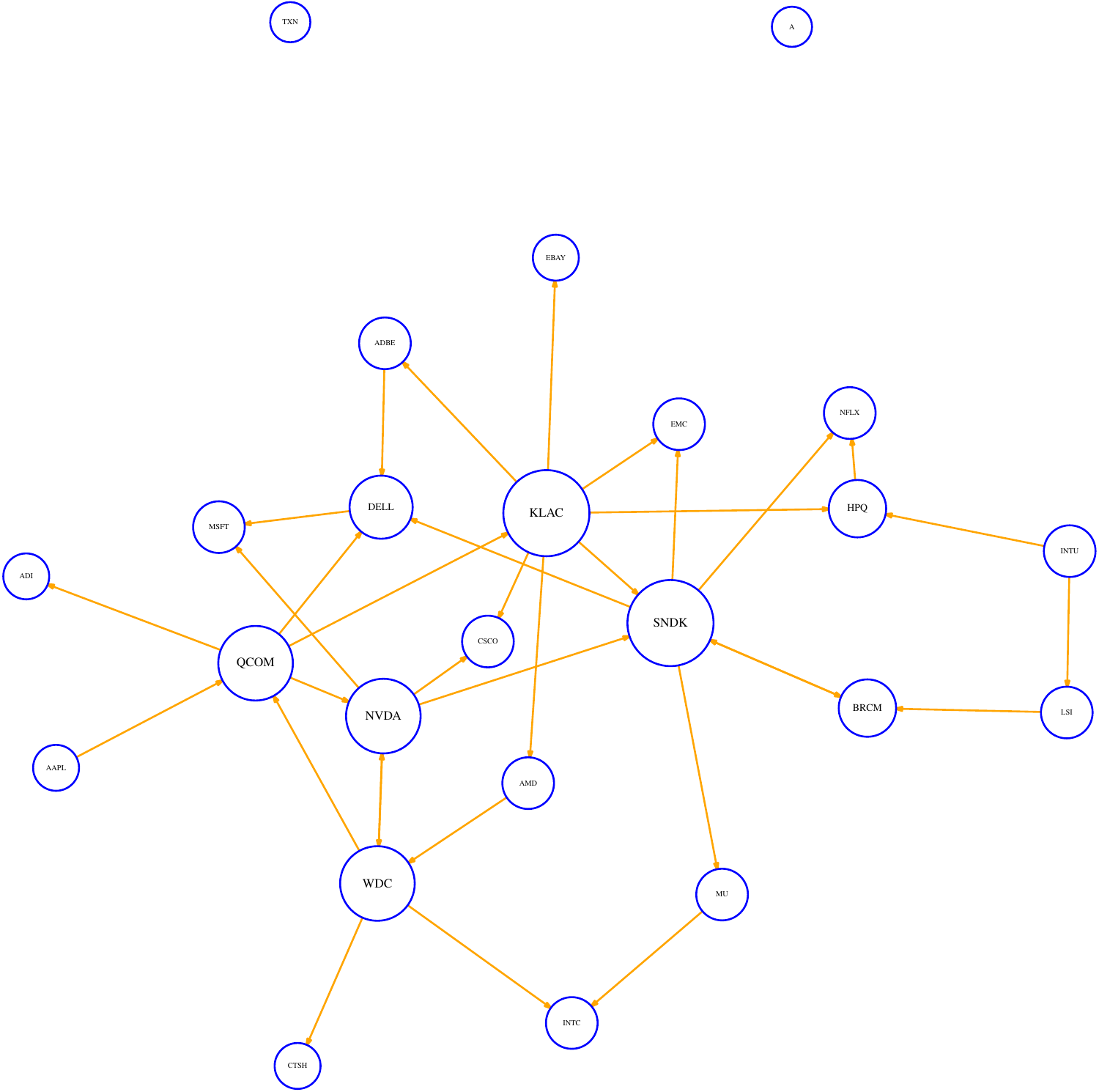}
		\\
		{\footnotesize (a) PISTA}
		& {\footnotesize (b) glmnet}
	\end{tabular}
	\caption{Comparing the causal interpretation of the results obtained by PISTA and glmnet side by side. The demonstrated companies' labels and the full names of the companies are listed in Table \ref{tab::tickers}.}
	\label{fig::stock}
\end{figure}

Another interesting interpretation that can be seen from Figure \ref{fig::stock} is that the glmnet does not recover any causal influence on the price of A among the companies in Table \ref{tab::tickers}, while PISTA interprets a causal relationship between A and HPQ. A closer look into the company history reveals that Agilent is a spinoff from the Hewlett-Packard in 1999 \citep{agilent}, and the prices of the two companies share similar trend even until today.

\begin{table}[!t]
	\centering
	\begin{tabular}{|c|c||c|c|}
		\hline 	Label & Company Name & Label & Company Name \\\hline
		ADBE & Adobe & AMD & Advanced Micro Devices\\\hline
		A & Agilent Technologies & ADI & Analog Devices \\\hline
		AAPL & Apple & BRCM & Broadcom \\\hline
		CSCO & Cisco Systems & CTSH & Cognizant \\\hline
		DELL & Dell & EBAY & eBay \\\hline
		EMC & Dell EMC & HPQ & Hewlett Packard Enterprise \\\hline INTC & Intel & INTU & Intuit \\\hline 
		KLAC & KLA-Tencor & LSI & LSI Corporation \\\hline
		MU & Micron Technology & MSFT & Microsoft Corporation \\\hline NFLX & Netflix & NVDA & Nvidia \\\hline QCOM & Qualcomm & SNDK & SanDisk \\\hline  TXN & Texas Instruments & WDC & Western Digital
		\\\hline
	\end{tabular}
	\caption{Labels and corresponding company names used in Figure \ref{fig::stock}.}
	\label{tab::tickers}
\end{table}

\section{Conclusions}\label{sec:conclusion}

In this paper, we studied the sparse additive models for discrete time series (TS-SpAMs) in high dimensions. We proposed an estimator that fits this model by exploiting the basis expansion and nonconvex regularization techniques. We provided both theoretical guarantees and a computationally efficient algorithm, PISTA, that tames the nonconvexity while guaranteeing a linear convergence towards the desired sparse solution with high probability. Our result has direct implications on detecting nonlinear causality in large time series datasets, and we compared the result of recovering causal relationships between PISTA and glmnet on a stock market dataset.





\bibliographystyle{ims}
{\small
	\setlength{\bibsep}{1.8pt}
	\bibliography{spad_noncvx}
}

\newpage\onecolumn

\section{Appendix: Proofs of the Major Theoretical Results}\label{sec:appendix}

For convenience of proving the main theorem, the following technical lemmas are needed.

\subsection{ Some Basic Definitions and Lemmas}


\begin{definition}
	For a loss function $\calL(\beta)$ that is twice differentiable, and a given $s$, define 
	\begin{align}\label{eq:rho}
	\rho_+(s) = \sup_{\|z\|_{0,2}\le s} {z^\top\nabla^2\Lc(\beta)z \over \|z\|_2^2}, ~~
	\rho_-(s) = \inf_{\|z\|_{0,2}\le s} {z^\top\nabla^2\Lc(\beta)z \over \|z\|_2^2}.
	\end{align}
\end{definition}
\noindent Under this definition, $\rho_+(s)$ and $\rho_-(s)$ are the largest and smallest $s$-block-sparse eigenvalues of $\nabla^2\Lc(\beta)$, respectively.
\begin{definition}
	For all $\beta,\beta'\in\reals^{pq}$,
	we say that the function $h(\beta)$ is $a$-strongly convex ($a$-SC) if there is a positive real number $a$, such that
	\begin{align*}
	h(\beta') - h(\beta) - (\beta' -\beta)^\top\nabla h(\beta) \ge {a\over 2}\|\beta'-\beta\|_2^2.
	\end{align*}
	We say that $h(\beta)$ is $b$-smooth ($b$-S) if there is a positive real number $b$, such that
	\begin{align*}
	h(\beta') - h(\beta) - (\beta' -\beta)^\top\nabla h(\beta) \le {b\over 2}\|\beta'-\beta\|_2^2.
	\end{align*}
\end{definition}
\noindent When the above equations only hold in a specific subset of $\reals^{pq}$, we say that $h(\beta)$ is $a$-restricted strongly convex ($a$-RSC) or $b$-restricted smooth ($b$-RS).



\begin{lemma}\label{lem:rsc}
	Suppose that $0\leq C_{\min}'q^{-1}\leq\rho_{-}(s)\leq\rho_{+}(s)\leq C_{\max}'q^{-1}<\infty$. Then, for any $\beta,\beta'\in \reals^{pq}$ such that $|\{j : \|\beta_j\|_2\ne 0 ~\text{or}~ \|\beta'_j\|_2\ne 0  \} | \le s\leq C_\calP$, we have
	\begin{enumerate}
		\item[a)]  $L(\beta)$ is ${\rho_-(s)\over 2}$-RSC and ${\rho_+(s)\over 2}$-RS.
		\begin{align*}
		{\rho_-(s)\over 2} \|\beta'-\beta\|_2^2 
		\le& L(\beta') - L(\beta) - (\beta' -\beta)^\top\nabla L(\beta)
		\le {\rho_+(s)\over 2} \|\beta'-\beta\|_2^2.
		\end{align*}
		\item[b)] $\tilde{L}(\beta)$ is ${\rhot_-(s)\over 2}$-RSC and ${\rho_+(s)\over 2}$-RS.
		\begin{align*}
		{\rhot_-(s)\over 2} \|\beta'-\beta\|_2^2 \le \tilde{L}_\lambda(\beta') - \tilde{L}_\lambda(\beta) - (\beta' -\beta)^\top\nabla\tilde{L}_\lambda(\beta) \le {\rho_+(s)\over 2} \|\beta'-\beta\|_2^2
		\end{align*}
		with $\rhot_-(s) = \rho_-(s) - \eta_-$.
		\item[c)] For all $\xi\in\partial \|\beta\|_{1,2}$, $
		\Fc_\lambda(\beta)$ is ${\rhot_-(s)\over 2}$-RSC.
		\begin{align*}
		{\rhot_-(s)\over 2} \|\beta'-\beta\|_2^2 \le \Fc_\lambda(\beta') - \Fc_\lambda(\beta) - (\beta' -\beta)^\top(\nabla\tilde{L}_\lambda(\beta) + \lambda\xi).
		\end{align*}
	\end{enumerate}
	
\end{lemma}

\begin{lemma}\label{lem:prob}
	Suppose that the blocks of the design matrix $Z$, which we defined below equation (\ref{eq:beta_sam}), satisfy the restricted eigenvalue condition specified in Lemma \ref{lm:rip}. Suppose that the noise $\epsilon$ is sub-Gaussian and has independent entries with $\Expect\epsilon_i=0$ and $\Var(\epsilon_i)<\infty$. Then, for $q=\calO(n^{1/(2d+1)})$, there exists $C_1$, $C_2$ such that
	\begin{align*}
	\Prob\left[\left\|\frac{1}{n}Z^\top(y-Z\beta^*)\right\|_{\infty,2}\leq\frac{C_1}{\sqrt{n}}+\frac{C_2\sqrt{\log p}}{n}\right]\geq 1-p^{-1}.
	\end{align*}
	
\end{lemma}

\begin{lemma}\label{lem:lambda}
	Under the same condition as in Lemma \ref{lem:prob}, let the regularization parameter $\lambda$ satisfy
	\[\lambda=8\sigma\max\{C_1,C_2\} \left(\frac{1}{\sqrt{n}}+\frac{\sqrt{\log p}}{n}\right).\]
	Then 
	$$\Pb(\|\nabla\Lc(\beta^*) \|_{\infty,2}\leq\lambda/4) \ge 1-p^{-1}.$$
\end{lemma}
In Lemma \ref{lem:lambda}, the constants $C_1$ and $C_2$ are the same constants as in Lemma \ref{lem:prob}.

\begin{lemma}\label{lem:E1}
	Let $\lambda$ be chosen the same way as in Lemma \ref{lem:lambda}. Suppose that the relationship (\ref{strong-signal}) and the event
	\begin{align*}
	\Ec_1 = \left\{ {1\over n}\left\|Z^\top(y-Z\beta^*)\right\|_{\infty,2} \le \lambda/4 \right\}
	\end{align*}
	hold, then we have
	\begin{align*}
	{1\over n} Z_S^{\top}(y-Z\betah^o) + \left[\nabla\Wc_{\lambda}(\betah^o) + \lambda \nabla \|\betah^o\|_{1,2}\right]_S = 0.
	\end{align*}
\end{lemma}

\begin{lemma}\label{lem:E2}
	Let $\lambda$ be chosen the same way as before, and suppose that the event
	\begin{align*}
	\Ec_2 = \left\{ {1\over n}\left\|U^\Sbar(y-Z\beta^*)\right\|_{\infty,2} \le \lambda/4 \right\}
	\end{align*}
	hold, where 
	\[U^{\Sbar} = Z_{\Sbar}^{\top}(I-Z_{S}(Z_S^{\top}Z_S)^{-1}Z_S^\top).\]	
	Then there exists $\xih^o_\Sbar\in [\partial \|\betah^o\|_{1,2}]_\Sbar$ such that
	\begin{align*}
	{1\over n} Z_{\Sbar}^{\top} (y-Z\betah^o) + [\nabla W_{\lambda}(\betah^o)]_\Sbar + \lambda \xih^o_\Sbar = 0.
	\end{align*}
\end{lemma}

\begin{lemma}\label{lem::variance}
	Suppose $X_1,\ldots,X_n$ are $n$ random variables with finite variance. Then
	\$
	\var\left(\sum_{i=1}^{n}X_i\right)\leq n\sum_{i=1}^{n}\var(X_i).
	\$
\end{lemma}

\subsection{Proof of Lemma \ref{lm:approx}}

\begin{proof}
	
	We wish to bound the following quantity:
	\begin{align*}
	\sup_{f_j\in\calF}\left\|f_{j}-\tilde{f}_{j}\right\|_2
	\end{align*}
	for some carefully chosen $\tilde{f}_j$.
	By Assumption (A6), there exists a $\tilde{f}_{j}^*\in\calF_{jn}$ such that
	\begin{align}\label{eq::501}
	\left\|f_{j}-\tilde{f}_{j}^*\right\|_2=\calO(q^{-d}).
	\end{align}
	Let $\tilde{f}_{j}\in\calF_{jn}^0$ be the centered version of $\tilde{f}_{j}^*$:
	\begin{align*}
	\tilde{f}_{j}=\tilde{f}_{j}^*-\frac{1}{n}\sum_{t=1}^{n}\tilde{f}_{j}^*(X_j[t])=\tilde{f}_j^*-P_n\tilde{f}_j^*,
	\end{align*}
	with $\{X_j([t])\}_{t=1}^{n}$ being a strong mixing sequence. Then, we have, upon invoking the triangle inequality, that
	\begin{align}\label{eq::502}
	\left\|f_{j}-\tilde{f}_{j}\right\|_2&=\left\|f_{j}-\tilde{f}_{j}^*+P_n\tilde{f}_{j}^*\right\|_2\leq\left\|f_{j}-\tilde{f}_{j}^*\right\|_2+\left\|P_n\tilde{f}_{j}^*\right\|_2=\left\|f_{j}-\tilde{f}_{j}^*\right\|_2+\sqrt{b-a}\left|P_n\tilde{f}_{j}^*\right|,
	\end{align}
	where $[a,b]$ is the support of $f_j(\cdot)$ for all $j$s.
	The first term has already been bounded by $\calO(q^{-d})$, and we now bound $|P_n\tilde{f}_{j}^*|$. Notice that, $Pf_j=0$. Therefore
	\begin{align}\label{eq::503}
	P_n\tilde{f}_{j}^*=(P_n-P)\tilde{f}_{j}^*+P(\tilde{f}_{j}^*-f_{j}),
	\end{align}
	where
	\begin{align*}
	P\tilde{f}_{j}^*=\Expect\left[\frac{1}{n}\sum_{t=1}^{n}\tilde{f}_{j}^*(X_j[t])\right],
	\end{align*}
	and
	\begin{align*}
	Pf_{j}=\Expect\left[\frac{1}{n}\sum_{t=1}^{n}f_{j}(X_j[t])\right].
	\end{align*}
	By Assumption (A6), 
	\begin{align}\label{eq::504}
	\left|P(\tilde{f}_{j}^*-f_{j})\right|=\calO(q^{-d}).
	\end{align}
	Hence, the only thing left to be shown is
	\begin{align}\label{eq::505}
	\sup_{\tilde{f}_{j}^*\in\calF_{jn}}\left|(P_n-P)\tilde{f}_{j}^*\right|=\calO(n^{-1/2}q^{1/2}),
	\end{align}
	which follows directly from the standard trick of chaining (see Chapter 5 of \cite{van2014probability}).
	More specifically, we invoke the following tail chaining inequality (Theorem 5.29 of \cite{van2014probability}):
	\begin{align*}
	\Prob\left[\sup_{\tilde{f}_{j}^*\in\calF_{jn}}\left|(P_n-P)\tilde{f}_{j}^*\right|\geq C_1\int_0^\infty\sqrt{\log N_{[\cdot]}(\epsilon,\calF_{jn},L_2(P))}d\epsilon+x\right]\leq C_1\exp\left\{-\frac{x^2}{C_4{\diam}^2(\calF_{jn})}\right\},
	\end{align*}
	where ${\diam}(\calF_{jn})$ is the diameter of $\calF_{jn}$ under the metric $L_2(P)$.
	By the result on page 597 of \cite{shen1994convergence}, the bracketing number of $\calF_{jn}$ is given by
	\begin{align*}
	\log N_{[\cdot]}(\epsilon,\calF_{jn},L_2(P))\leq C_2 q\log \epsilon^{-1}
	\end{align*}
	for some constant $C_2>0$ and $\epsilon\in[0,1]$, and is 0 otherwise. Therefore, by choosing $C_1=n^{-1/2}$ and $x=n^{-1/2}q^{1/2}$, and noticing that ${\diam}(\calF_{jn})<\infty$, we have
	\begin{align*}
	\Prob\left[\sup_{\hat{f}_{j}^*\in\calF_{jn}}\left|(P_n-P)\tilde{f}_{j}^*\right|\geq C_3n^{-1/2}q^{1/2}\right]\leq n^{-1/2}.
	\end{align*}
	As $n\to\infty$, the probability converges to 0. Hence \eqref{eq::505} is proved. The lemma follows by combining the equations \eqref{eq::501}, \eqref{eq::502}, \eqref{eq::503}, \eqref{eq::504}, and \eqref{eq::505}.
	
\end{proof}

\subsection{Proof of Lemma \ref{lm:rip}}

\begin{proof}
	We prove the lemma in two steps. In the first step, we show that $Z_i^{\top}Z_i/n$ concentrates tightly around its expectation in terms of the operator norm. Hence, upon grasping the spectrum of the $\Expect[Z_i^{\top}Z_i/n]$, the result follows easily. In the second step, we invoke the union bound to obtain a uniform concentration bound. As we shall see, the maximum over $i$ has little effect on the overall rate of concentration because the tail probability decays exponentially with $n$.
	
	{\bfseries\noindent Step 1: Concentration of $Z_i^\top Z_i/n$.} 
	
	{\bfseries\noindent Simplifying the representation of $\|Z_i\|_{\op}^2$.} 
	Suppose $v\in\reals^q$ is a vector of unit norm: $\|v\|_2=1$, or $v\in\mathbb{S}^{q-1}$ where $\mathbb{S}^{q-1}$ denotes the $(q-1)$-dimensional unit sphere. Then the operator norm of $Z_i$ can be written as
	\begin{align*}
	\|Z_i\|_{\op}^2=\sup_{v\in\mathbb{S}^{q-1}}v^\top Z_i^\top Z_i v=\sup_{v\in\mathbb{S}^{q-1}}\sum_{t=1}^{n}\lbrac\sum_{j=1}^{q}\psi_{ij}(X_i[t])v_j\rbrac^2.
	\end{align*}
	Since we consider both the maximum and minimum eigenvalues of $Z_i$, we define
	\$
	\|Z_i\|_{\op}^2(v)=\sum_{t=1}^{n}\lbrac\sum_{j=1}^{q}\psi_{ij}(X_i[t])v_j\rbrac^2,
	\$
	and therefore the maximum and minimum eigenvalues are determined by the maximum and minimum values of $\|Z_i\|_{\op}^2(v)$. Recall that $\psi_{ij}$ is the centered version of $\phi_{j}$ using the samples of $X_i[t]$. Thus, with simple manipulations, we further have
	\$
	\|Z_i\|_{\op}^2(v)=\sum_{t=1}^{n}\left[\sum_{j=1}^{q}\left(\phi_{j}(X_i[t])-\frac{1}{n}\sum_{\tau=1}^{n}\phi_{j}(X_i[\tau])\right)v_j\right]^2.
	\$
	Denote
	\begin{align*}
	s(x)=\sum_{j=1}^{q}\phi_{j}(x)v_j.
	\end{align*}
	Then,
	\$
	\|Z_i\|_{\op}^2(v)&=\sum_{t=1}^{n}\left[s(X_i[t])-\frac{1}{n}\sum_{\tau=1}^{n}s(X_i[\tau])\right]^2=\sum_{t=1}^{n}s^2(X_i[t])-\frac{1}{n}\left[\sum_{t=1}^{n}s(X_i[t])\right]^2\\&=n\left\{\frac{1}{n}\sum_{t=1}^{n}s^2(X_i[t])-\left[\frac{1}{n}\sum_{t=1}^{n}s(X_i[t])\right]^2\right\}.
	\$
	
	{\bfseries\noindent Concentration of $\|Z_i\|_{\op}^2(v)$.}	We now wish to show concentration of $\|Z_i\|_{\op}^2(v)$ to $\Expect\|Z_i\|_{\op}^2(v)$ for any fixed choice of $v$. Define
	\begin{align*}
	S_n(v)=\|Z_i\|_{\op}^2(v)-\Expect\|Z_i\|_{\op}^2(v).
	\end{align*}
	Then,
	\$
	S_n(v)&=\left\{\sum_{t=1}^{n}s^2(X_i[t])-\Expect\left[\sum_{t=1}^{n}s^2(X_i[t])\right]\right\}-\frac{1}{n}\left\{\left[\sum_{t=1}^{n}s(X_i[t])\right]^2-\Expect\left[\sum_{t=1}^{n}s(X_i[t])\right]^2\right\}\\&\triangleq S_{1,n}(v)-\frac{1}{n}S_{2,n}(v),
	\$
	where $S_{1,n}(v)$ and $S_{2,n}(v)$ represent the summation of the variance and the variance of the summation. By Lemma \ref{lem::variance}, $|n^{-1}S_{2,n}(v)|\leq |S_{1,n}(v)|$. Hence, $|S_n(v)|\leq 2|S_{1,n}(v)|$. 
	
	 Since $\{X_i[t]\}_{t=1}^{\infty}$ is a strong mixing sequence, $\{s(X_i[t])\}_{t=1}^{\infty}$ is also strong mixing. Therefore, from Lemma \ref{lm:Bernstein}, we know that for some constant $C_{\tilde{\mu}}$,
	\begin{align*}
	\Prob\lbrac\frac{|S_{1,n}(v)|}{n}\geq \delta\rbrac&\leq 2\exp\lbrac-\frac{C_{\tilde{\mu}}n\delta^2}{M^2+M\delta\log n\log\log n}\rbrac.
	\end{align*}
	
	{\bfseries\noindent Step 2: Bounding the expectation.} By assumption (A6), we have the empirical variance and the empirical second moment of $s(X_i[t])$ are on the same order, and therefore it suffices to consider the expectation of the empirical second moment, the maximum of which over $i\in \{1,\ldots, p\}$ is
	\begin{align*}
	\max_{i\in[p]}\Expect\left[\lbrac\sum_{j=1}^{q}\phi_{j}(X_i[t])v_j\rbrac^2\right]&=\max_{i\in[p]}\int_{a}^{b}s^2(x)dQ_{i,t}(x)\leq q_{\max}\int_a^bs^2(x)dx\notag\\&\leq q_{\max}\sum_{j=1}^{q}v_j^2(\xi_j-\xi_{j-l})\leq C_1q^{-1}
	\end{align*} for some $C_1>0$, and
	\begin{align*}
	\min_{i\in[p]}\Expect\left[\lbrac\sum_{j=1}^{q}\phi_j(X_i[t])v_j\rbrac^2\right]&=\min_{i\in[p]}\int_{a}^{b}s^2(x)dQ_{i,t}(x)\geq q_{\min}\int_a^bs^2(x)dx\notag\\&\geq q_{\min}\sum_{j=1}^{q}v_j^2(\xi_j-\xi_{j-l})\geq C_2q^{-1}	
	\end{align*}
	for some $C_2>0$. Hence, for any $i\in[p]$,
	\begin{align*}
	\Prob\left[\frac{\|Z_iv\|^2}{n}\geq \delta+C_1q^{-1}\right]\leq 2\exp\lbrac-\frac{C_{\tilde{\mu}}n(\delta+C_1q^{-1})^2}{M^2+M(\delta+C_1q^{-1})\log n\log\log n}\rbrac.
	\end{align*}
	
	{\bfseries\noindent Step 3: Invoking the union bound.}	Let $\calV$ be an $\epsilon$-net for the sphere $\mathbb{S}^{q-1}$, then for any $v\in\mathbb{S}^{q-1}$, there exists a $\tilde{v}\in\calV$ such that $\|v-\tilde{v}\|\leq\epsilon$. Hence, by letting $v$ be the vector such that $\|Z_iv\|=\|Z_i\|_{\op}$, we have \[\|Z_i\|_{\op}=\|Z_iv\|\leq\|Z_i(\tilde{v}-v)\|+\|Z_i\tilde{v}\|\leq\epsilon\|Z_i\|_{\op}+\|Z_i\tilde{v}\|.\] Hence,
	\begin{align*}
	\|Z_i\|_{\op}\leq\frac{1}{1-\epsilon}\max_{v\in\calV}\|Z_iv\|,
	\end{align*}
	which implies that
	\begin{align*}
	\Prob\left[\max_{i\in[p]}\sup_{v\in\mathbb{S}^{q-1}}\frac{\|Z_iv\|^2}{n}\geq \delta+C_1q^{-1}\right]&\leq p\Prob\left[\max_{v\in\calV}\frac{\|Z_iv\|^2}{n}\geq (\delta+C_1q^{-1})(1-\epsilon)^2\right]\leq 2|\calV|P_\epsilon,
	\end{align*}
	where \[P_\epsilon=\exp\lbrac\log p-\frac{C_{\tilde{\mu}}n(\delta+C_1q^{-1})^2(1-\epsilon)^4}{M^2+M(\delta+C_1q^{-1})(1-\epsilon)^2\log n\log\log n}\rbrac.\]
	If we use a maximal $\epsilon$-packing of $\mathbb{S}^{q-1}$ as the $\epsilon$-net, then \[|\calV|\leq(1+2\epsilon^{-1})^{q}.\] Hence,
	\begin{align*}
	\Prob\left[\max_{i\in[p]}\frac{\|Z_i\|_{\op}^2}{n}\geq \delta+C_1q^{-1}\right]\leq\inf_{\epsilon>0}\left\{2\lbrac 1+\frac{2}{\epsilon}\rbrac^qP_\epsilon\right\}.
	\end{align*}
	Let $\delta=C_3q^{-1}$. Since in a high dimensional setting, $\log p$ is a very small polynomial of $n$, we thus have, when $q\asymp n^{-\nu}$ with $\nu\in(0,1/2)$, $\|Z_i\|_{\op}^2\leq C_{\max}nq^{-1}$ with high probability.

	The lower bound can be obtained in the same fashion. Hence, with high probability, $C_{\min}q^{-1}\leq\rho_{\min}(U_i)\leq\rho_{\max}(U_i)\leq C_{\max}q^{-1}$.
\end{proof}

\subsection{Proof of Lemma \ref{lm:rse}}

The proof for this Lemma coincides with the proof of Lemma 3 in \cite{HuangHW10}.

\subsection{Proof of Lemma \ref{lem:rsc}}
\begin{proof}
	First, $|\{j : \|\beta_j\|_2\ne 0 ~\text{or}~ \|\beta'_j\|_2\ne 0  \} | \le s$ implies
	$\|\beta-\beta'\|_{0,2}\le s$. 
	\begin{enumerate}
		\item[a)]  Since $L(\beta)$ is twice differentiable, by the mean value theorem, there is $\mu\in(0,1)$ and $\betat=(1-\mu)\beta' + \mu\beta$, such that
		\begin{align}\label{eq:rsc:1}
		L(\beta') - L(\beta) - (\beta'-\beta)^\top\nabla L(\beta) = {1\over 2}(\beta'-\beta)^\top\nabla^2 L(\betat)(\beta'-\beta).
		\end{align}
		According to the definition of $\rho_-(s)$ and $\rho_+(s)$ in \eqref{eq:rho}, we have
		\begin{align}\label{eq:rsc:2}
		{\rho_-(s)\over 2} \|\beta'-\beta\|_2^2 \le
		{1\over 2}(\beta'-\beta)^\top\nabla^2 L(\betat)(\beta'-\beta)
		\le {\rho_+(s)\over 2} \|\beta'-\beta\|_2^2 .
		\end{align}
		Then \eqref{eq:rsc:1} and \eqref{eq:rsc:2} indicate
		\begin{align}\label{eq:rsc:3}
		{\rho_-(s)\over 2} \|\beta'-\beta\|_2^2 
		\le L(\beta') - L(\beta) - (\beta'-\beta)^\top\nabla L(\beta) 
		\le {\rho_+(s)\over 2} \|\beta'-\beta\|_2^2 .
		\end{align}
		\item[b)] By Regularity d) and the concavity of $\Wc_\lambda(\beta)$, we have
		\begin{align*}
		-{\eta_-\over 2}\|\beta'-\beta\|_2^2 \le \Wc_\lambda(\beta') - \Wc_\lambda(\beta) - (\beta'-\beta)^\top\nabla\Wc_\lambda(\beta) \le 0
		\end{align*}
		which, together with \eqref{eq:rsc:3}, implies
		\begin{align}\label{eq:rsc:4}
		{\rhot_-(s)\over 2} \|\beta'-\beta\|_2^2 \le \tilde{L}_\lambda(\beta') - \tilde{L}_\lambda(\beta) - (\beta' -\beta)^\top\nabla\tilde{L}_\lambda(\beta) \le {\rho_+(s)\over 2} \|\beta'-\beta\|_2^2,
		\end{align}
		where $\rhot_-(s) = \rho_-(s) - \eta_-$.
		\item[c)] For all $\xi\in\partial \|\beta\|_{1,2}$, since $\|\beta\|_{1,2}$ is convex w.r.t 
		$\beta$, we have $\|\beta'\|_{1,2} \ge \|\beta\|_{1,2} +(\beta'-\beta)^\top\xi$, which, after being plugged into \eqref{eq:rsc:4}, gives 
		\begin{align*}
		{\rhot_-(s)\over 2} \|\beta'-\beta\|_2^2 \le \Fc_\lambda(\beta') - \Fc_\lambda(\beta) - (\beta' -\beta)^\top(\nabla\tilde{L}_\lambda(\beta) + \lambda\xi).
		\end{align*}
	\end{enumerate}
\end{proof}

\subsection{Proof of Lemma \ref{lem:prob}}
\begin{proof}
	Consider $y=(X_1[2],...,X_1[n])^{\top}$. Recall that $\|\beta\|_{\infty,2}=\max_{j=1,...,p}\|\beta_j\|_2$. We hence have
	\begin{align*}
	\left\|\frac{Z^{\top}(y-Z\beta^*)}{n}\right\|_{\infty,2}=\max_{i=1,2,...,p}\left\|\frac{1}{n}Z_i^{\top}(y-Z\beta^*)\right\|_2.
	\end{align*}
	Since
	\begin{align*}
	y-Z\beta^*=\epsilon+\left[\begin{array}{c}\sum_{j=1}^{p}f_{j}(X_j[1])\\\sum_{j=1}^{p}f_{j}(X_j[2])\\\vdots\\\sum_{j=1}^{p}f_{j}(X_j[n-1])\end{array}\right]-\left[\begin{array}{c}\sum_{j=1}^{p}\hat{f}_{j}(X_j[1])\\\sum_{j=1}^{p}\hat{f}_{j}(X_j[2])\\\vdots\\\sum_{j=1}^{p}\hat{f}_{j}(X_j[n-1])\end{array}\right],
	\end{align*}
	where we have assumed that the true functions are $f_{j}(\cdot)$s, while $\hat{f}_{j}(\cdot)$s are the functions with truncated spline expansion, we readily see that $y-Z\beta^*$ consists of the noise in the model and the error caused by the truncated spline approximation. For simplicity, we denote the error caused by spline approximation as $E[t]$ for $t=1,...,n-1$, and let $E=(E[1],\ldots,E[t-1])^{\top}$. Thus, for each block $Z_i$,
	\begin{align*}
	\left\|\frac{1}{n}Z_i^{\top}(y-Z\beta^*)\right\|_2&=\left\|\frac{1}{n}Z_i^{\top}\epsilon+\frac{1}{n}Z_i^{\top}E\right\|_2\leq\left\|\frac{1}{n}Z_i^{\top}\epsilon\right\|_2+\left\|\frac{1}{n}Z_i^{\top}E\right\|_2.
	\end{align*}
	By Lemma \ref{lm:approx}, choose $q=\calO(n^{1/(2d+1)})$, we have \[\|f_{j}-\hat{f}_{j}\|_2=\calO(n^{-d/(2d+1)}).\] Hence, 
	\begin{align*}
	\left\|\frac{1}{n}Z_i^{\top}E\right\|_2&\leq\frac{\sqrt{q}}{n}\sum_{t=1}^{n-1}\phi_1(X_i[t])\sum_{j=1}^{p}(f_{j}-\hat{f}_{j})(X_j[t])\leq C\sqrt{q}\sum_{j\in\calP_1}(f_{j}-\hat{f}_{j})(X_i[t])\\&\leq C\sqrt{q}n^{-d/(2d+1)}\leq Cn^{-1/2},
	\end{align*}
	where we have used the fact that when $f_{j}(\cdot)\equiv 0$, it can be represented exactly by a truncated spline expansion with the corresponding coefficients being zeros.
	
	We now upper bound the other term, $\|\frac{1}{n}Z_i^{\top}\epsilon\|_2$. We do so by upper bounding $n^{-2}\epsilon^{\top}Z_iZ_i^{\top}\epsilon$. Fix $Z_i$. By assumption, we have $C_{\min}q^{-1}n\leq\rho_{\max}(Z_iZ_i^{\top})\leq C_{\max}q^{-1}n$. By Theorem 2.1 of \cite{2013arXiv1306.2872R},
	\begin{align*}
	\Prob[|\|Z_i^{\top}\epsilon\|_2-\|Z_i\|_{HS}|>\lambda]\leq \exp\lbrac-\frac{c\lambda^2}{K^4\|Z_i\|^2}\rbrac,
	\end{align*}
	for some constant $c$. Here, $K$ is the uniform upper bound for \[\|\epsilon_j\|_{\psi_2}=\sup_{k\geq 1}k^{-1/2}(\Expect|\epsilon_j|^k)^{1/k}.\]
	Here, $\epsilon_j$ is the $i$-th dimension of $\epsilon$, and $\epsilon=(\epsilon_i[1],\ldots,\epsilon_i[n])^{\top}$. 
	
	Since $\epsilon$ is subgaussian, $\epsilon_j$ is subgaussian as well. Hence $K<\infty$. By assumption, we have $\Expect \epsilon\epsilon^{\top}=\sigma^2I$, with $I$ being the identity matrix. We also have the non-zero eigenvalues of $Z_iZ_i^{\top}$ satisfying the restricted eigenvalue condition. Since the non-zero eigenvalues of $Z_iZ_i^{\top}$ coincide with $Z_i^{\top}Z_i$, we have, when $n>q$, $\tr(Z_iZ_i^{\top})=\tr(Z_i^{\top}Z_i)=q\times (C_inq^{-1})=C_in$ for some constant $C_i$. Hence \[\|Z_i\|_{HS}=\sqrt{\Expect\epsilon^{\top}Z_iZ_i^{\top}\epsilon}=\sqrt{\tr(Z_iZ_i^{\top})\Expect[\epsilon^{\top}\epsilon]}=C_i\sqrt{n}\] for some constant $C_i$. Since $Z_i$ has bounded entries, the operator norm of $Z_i$ is also upper bounded.
	
	Let $C_1=\max_{i\in[p]} C_i$. We can conclude, by union bound, that
	\begin{align*}
	\Prob\left[\sup_{i\in[p]}\frac{1}{n}\|Z_i^{\top}\epsilon\|_2>\frac{C_1}{\sqrt{n}}+\lambda\right]\leq p\exp(-cn^2\lambda^2)
	\end{align*}
	for some constant $c$.
	Set $\lambda=\sqrt{\log p/c}/n$, we get
	\begin{align*}
	\Prob\left[\sup_i\|Z_i^{\top}\epsilon\|_2>\frac{C_1}{\sqrt{n}}+\frac{\sqrt{\log p/c}}{n}\right]\leq p\exp(-\log p)=p^{-1}.
	\end{align*}
	Thus, combining with the previous argument on $\|\frac{1}{n}Z_i^{\top}E\|_2$, we have
	\begin{align*}
	\Prob\left[\left\|\frac{1}{n}Z^{\top}(y-Z\beta^*)\right\|_{\infty,2}\leq\frac{C_1}{\sqrt{n}}+\frac{C_2\sqrt{\log p}}{n}\right]\geq 1-p^{-1}
	\end{align*}
	for some constants $C_1$ and $C_2$.
\end{proof}

\subsection{Proof of Lemma \ref{lem:lambda}}
\begin{proof}
	We have 
	\begin{equation*}
	\|\nabla\Lc(\beta^*)\|_{\infty,2} =\left\|{1\over n} Z^{\top}(y-Z\beta^*)\right\|_{\infty,2}.
	\end{equation*}
	The claim immediately follows by combining the
	choice of $\lambda$ and Lemma \ref{lem:prob}.
\end{proof}

\subsection{Proof of Lemma \ref{lem:E1}}
\begin{proof}
	Lemma \ref{lm:rse} implies that $\rho_-(s^*) > C_{\min}'q^{-1} > 0 $, so the equation (\ref{eq:oracle}) has a strongly convex objective with a unique solution:
	\begin{align*}
	\betah_S^o=(Z_S^{\top}Z_S)^{-1}Z_S^{\top}y.
	\end{align*}
	Given that the event
	\begin{align*}
	\Ec_1 = \left\{ {1\over n}\|Z^{\top}(y-Z\beta^*)\|_{\infty,2} \le \lambda/4 \right\}
	\end{align*}
	holds, we have 
	\begin{align*}
	\|\betah_S^o - \beta_S^* \|_{\infty, 2} &
	= \|(Z_S^{\top}Z_S)^{-1}Z_S^{\top}y - (Z_S^{\top}Z_S)^{-1}Z_S^{\top}Z_S\beta_S^*\|_{\infty,2} \\&
	= \|(Z_S^{\top}Z_S)^{-1}Z_S^{\top}(y - Z_S\beta_S^*)\|_{\infty,2}\\&
	\le \max_{g\in S} {1\over \rho_-(s^*)} \left\|{1\over n}Z_g^{\top}(y-Z\beta^*)\right\|_{2}\\&
	= {1\over n\rho_-(s^*)} \|Z_S^{\top}(y-Z\beta^*)\|_{\infty, 2}\\&
	\le {\lambda\over 4\rho_-(s^*)},
	\end{align*}
	which, together with the assumption placed by equation (\ref{strong-signal}), indicates that
	\begin{align*}
	\min_{g\in S}\|\betah_g^o\|_2 &
	= \min_{g\in S}\|\betah_g^o - \beta^*_g + \beta^*_g\|_2 \\&
	\ge \min_{g\in S} (\|\beta_g^*\|_2 - \|\betah_g^o - \beta^*_g\|_2) \\&
	\ge \min_{g\in S} \|\beta_g^*\|_2 - \max_{{g'}\in S}\|\betah_{g'}^o - \beta^*_{g'}\|_2\\&
	= \min_{g\in S} \|\beta_g^*\|_2 - \|\betah_S^o - \beta^*_S\|_{\infty, 2} \\&
	\ge \left(C_1\sigma q - 2(C_{\min}')^{-1}\sigma q \right) \frac{\lambda}{8\sigma} 
	\\& \ge  C q\lambda.
	\end{align*}
	The above relationship further implies the existence of a constant $\gamma$ defined in Regularity condition c), such that
	\begin{align*}
	\min_{g\in S}\|\betah_g^o\|_2  \ge \gamma q\lambda.
	\end{align*}
	Therefore, by Regularity condition c), we have 
	\begin{align*}
	\left[\nabla\Wc_{\lambda}(\betah^o) + \lambda \nabla \|\betah^o\|_{1,2}\right]_S = 0,
	\end{align*}
	which, together with the optimality condition of problem \eqref{eq:oracle},
	\begin{align*}
	{1\over n} Z_S^{\top}(y-Z_S\betah^o_S) = {1\over n} Z_S^{\top}(y-Z\betah^o) = 0,
	\end{align*}
	leads to
	\begin{align*}
	{1\over n} Z_S^{\top}(y-Z\betah^o) + \left[\nabla\Wc_{\lambda}(\betah^o) + \lambda \nabla \|\betah^o\|_{1,2}\right]_S = 0.
	\end{align*}
\end{proof}

\subsection{Proof of Lemma \ref{lem:E2}}
\begin{proof}
	First notice that
	\begin{align*}
	\|Z_\Sbar^{\top}(y-Z\betah^o)\|_{\infty,2} &
	= \|Z_\Sbar^{\top}(y-Z_S\betah^o_S)\|_{\infty,2}\\&
	=\|Z_\Sbar^{\top}[Z_S\beta_S^*+\epsilon+E-Z_S(Z_S^{\top}Z_S)^{-1}Z_S^{\top}(Z_S\beta_S^*+\epsilon+E)]\|_{\infty,2}\\&
	= \|Z_\Sbar^{\top}[I-Z_S(Z_S^{\top}Z_S)^{-1}Z_S^{\top}](\epsilon+E)\|_{\infty,2} \\&
	= \|U^\Sbar(\epsilon+E)\|_{\infty,2}.
	\end{align*}
	Since the event
	\begin{align*}
	\Ec_2 = \left\{ {1\over n}\|U^\Sbar(\epsilon+E)\|_{\infty,2} \le \lambda/4 \right\}
	\end{align*}
	holds, we have 
	\begin{align*}
	\left\|{1\over n}Z_\Sbar^{\top}(y-Z\betah^o)\right\|_{\infty,2} \le {\lambda\over 4},
	\end{align*}
	which immediately implies
	\begin{align*}
	\left\|{1\over n}Z_\Sbar^{\top}(y-Z\betah^o)\right\|_{\infty} \le {\lambda\over 4}.
	\end{align*}
	According to the Regularity condition c), we know that $[\nabla \Wc_\lambda(\betah^o)]_\Sbar=0$. 
	We also know that the elements of the subdifferential of $\|\betah^o_\Sbar\|_{1,2}$ lie in 
	$[-1,1]$. Therefore, there must be some $\xih^o_\Sbar\in [\partial \|\betah^o\|_{1,2}]_\Sbar$ such that
	\begin{align*}
	{1\over n} Z_{\Sbar}^{\top} (y-Z\betah^o) + [\nabla \Wc_\lambda(\betah^o)]_\Sbar + \lambda \xih^o_\Sbar = 0.
	\end{align*}
\end{proof}

\subsection{Proof of Lemma \ref{lem::variance}}

We use mathematical induction. First, the claim holds for $n=1$. Suppose it holds up to $n-1$ for some $n>1$. Then, with $\rho$ denoting the correlation coefficient, we have
\$
\var\left(\sum_{i=1}^{n}X_i\right)&=\var\left(\sum_{i=1}^{n-1}X_i\right)+\var(X_n)+2\rho \sqrt{\var\left(\sum_{i=1}^{n-1}X_i\right)\var(X_n)}\\&\leq(n-1)\sum_{i=1}^{n-1}\var(X_i)+\var(X_n) + 2\sqrt{\frac{\var\left(\sum_{i=1}^{n-1}X_i\right)}{n-1}\cdot(n-1)\var(X_n)}\\&\leq (n-1)\sum_{i=1}^{n-1}\var(X_i)+\var(X_n) + \frac{\var\left(\sum_{i=1}^{n-1}X_i\right)}{n-1}+(n-1)\var(X_n)\\&\leq n\sum_{i=1}^{n-1}\var(X_i)+n\var(X_n) = n\sum_{i=1}^{n}\var(X_i).
\$
\subsection{Proof of Theorem \ref{thm:recover}}

\begin{proof}
	
	
	We mainly prove that the support of the coefficient vector $\hat{\beta}_S^o$ can be recovered with high probability, from which point the rest of the proof coincides with the result in Corollary 2 in \cite{HuangHW10}.
	
	From Lemmas \ref{lem:E1} and \ref{lem:E2},  we know that, if both $\Ec_1$ and $\Ec_2$ hold, $\betah^o$ satisfies the KKT condition of \eqref{eq:beta_sam}, i.e., $\betah^o$ is a stationary point of \ref{eq:beta_sam}. Secondly, from Lemma \ref{lem:prob}, we know $\Pb(\Ec_1)\ge 1-p^{-1}$. Furthermore, since $\|I-Z_S(Z_S^\top Z_S)^{-1}Z_S^\top\|_2\leq 1$, we have, for all $g\in[p]$,
	\begin{align*}
	\|U^g\|_2 
	=& \|Z_g^{\top}[I-Z_S(Z_S^{\top}Z_S)^{-1}Z_S^{\top}]\|_{2} \\
	\le& \|Z_g\|_2 \|I-Z_S(Z_S^{\top}Z_S)^{-1}Z_S^{\top}\|_2
	\le \|Z_g\|_2 \leq \sqrt{nq^{-1}}.
	\end{align*}
	This implies that $U^g$ also satisfies the bounded eigenvalue condition as in Lemma \ref{lm:rip}, and hence the result of Lemma \ref{lem:prob} also applies to $U$, i.e., $\Pb(\Ec_2)\ge 1-p^{-1}$. Therefore,
	\begin{equation*}
	\Pb(\betah^o = \betah) = \Pb(\Ec_1\cap \Ec_2) = 1- \Pb(\bar\Ec_1 \cup \bar\Ec_2) \ge 1-p^{-1}.
	\end{equation*}
	On the other hand, we have
	\begin{align*}
	\norm{\hat{\beta}^{o}_{S} -\beta^*}_2&= \norm{(Z_{S}^\top Z_S)^{-1}Z_S^\top (Z_S\beta^*+\epsilon+E)-\beta^*}_2\\
	& = \norm{(Z_{S}^\top Z_S)^{-1}Z_S^\top\epsilon+(Z_{S}^\top Z_S)^{-1}Z_S^\top E}_2\\
	&\leq \norm{(Z_{S}^\top Z_S)^{-1}Z_S^\top\epsilon}_2+\norm{(Z_{S}^\top Z_S)^{-1}Z_S^\top E}_2.
	\end{align*}
	Using $\Lambda_{\max}(X)$ to denote the maximum eigenvalue of matrix $X$, we have
	\begin{align*}
	\norm{(Z_{S}^\top Z_S)^{-1}Z_S^\top}_2 &= \sqrt{\Lambda_{\max}((Z_{S}^\top Z_S)^{-1}Z_S^\top Z_S(Z_{S}^\top Z_S)^{-1}})\\
	&=\sqrt{1/\Lambda_{\min}(Z_{S}^\top Z_S)} \leq \sqrt{\frac{Cq}{n}}.
	\end{align*}
	Hence, 
	\begin{align}
	\norm{(Z_{S}^\top Z_S)^{-1}Z_S^\top\epsilon}_2 \leq C\frac{q}{\sqrt{n}}\quad\textrm{and}\quad\norm{(Z_{S}^\top Z_S)^{-1}Z_S^\top E}_2\leq \sqrt{\frac{Cq}{n}}\norm{E}_2.
	\end{align}
	
	Since $q=\calO(n^{-1/(2d+1)})$, we have, by an argument similar to that for bounding the approximation error in Lemma \ref{lem:prob}, $\|E\|_2\leq C'n^{-1/2}$. Hence, $$\|\hat{\beta}_S^o-\beta^*\|_2\leq C_1\frac{q}{\sqrt{n}}+C_2\frac{\sqrt{q}}{n}.$$ Since the true support has been recovered with high probability, the remainder of the proof coincides with the result in Corollary 2 in \cite{HuangHW10}, and \[\norm{\hat{f}_j-f_j}_2^2=\calO_P(n^{-2d/(2d+1)}).\]

\end{proof}

\subsection{Additional Simulation Results}\label{sec::moreplots}

Finally, we provide some additional simulation results in this section, which are listed as follows.

\begin{figure}
	\centering
	\begin{tabular}{cc}		
		\includegraphics[width=0.45\textwidth]{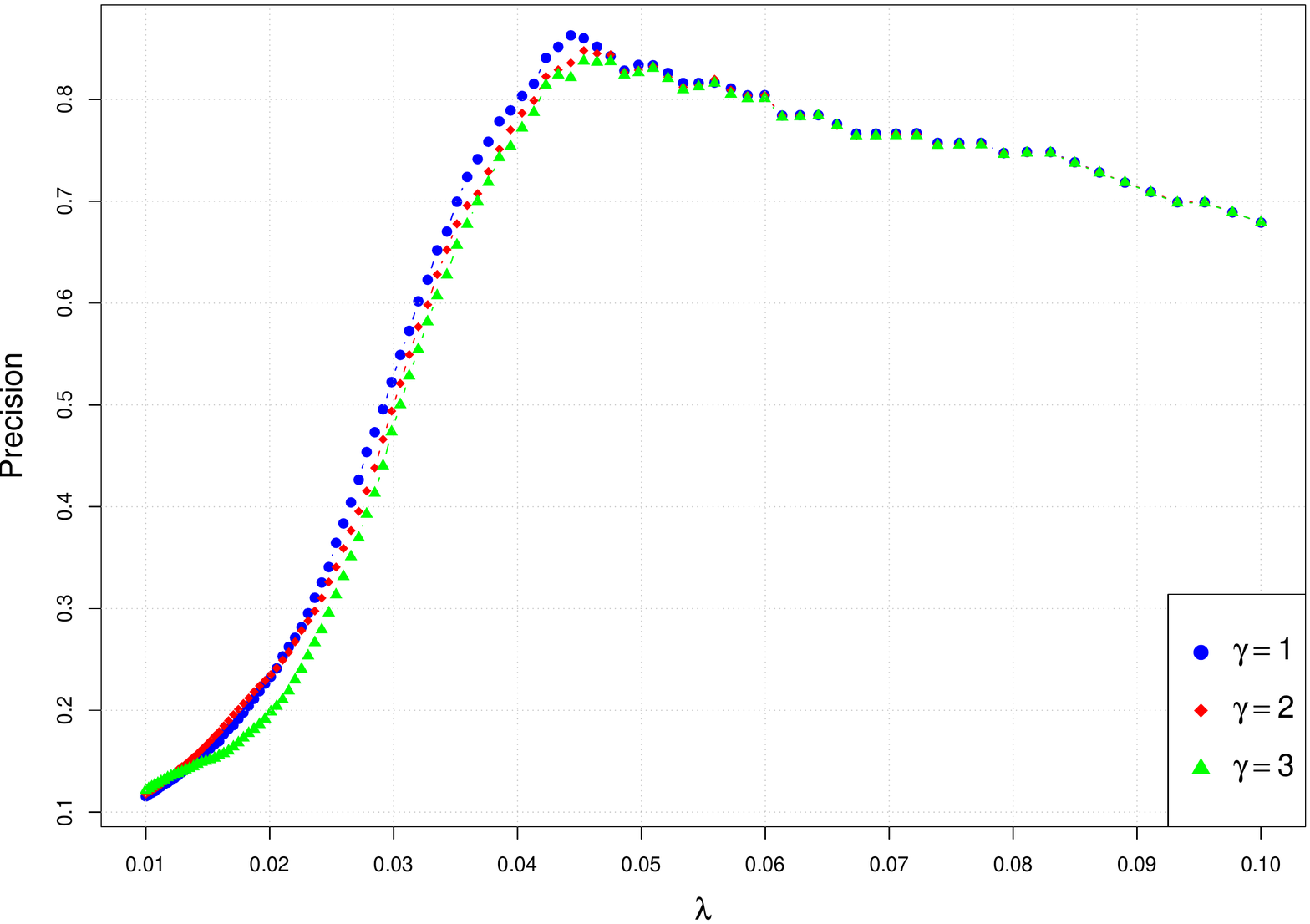}
		&\includegraphics[width=0.45\textwidth]{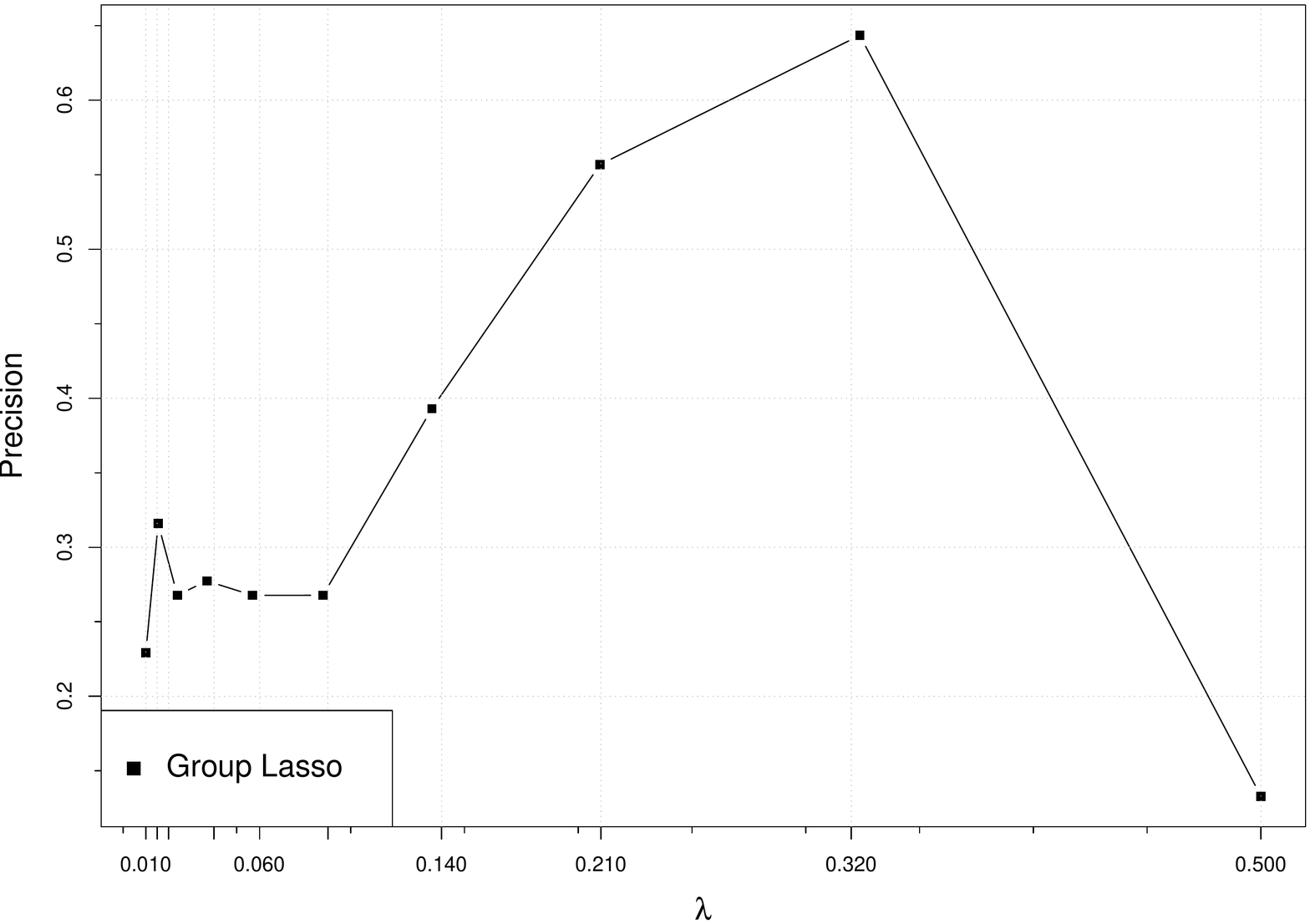}
		\\
		{\footnotesize (a) Precision of PISTA under different regularizations.}
		& {\footnotesize (b) Precision of group Lasso under different regularizations.}
		\\
		\includegraphics[width=0.45\textwidth]{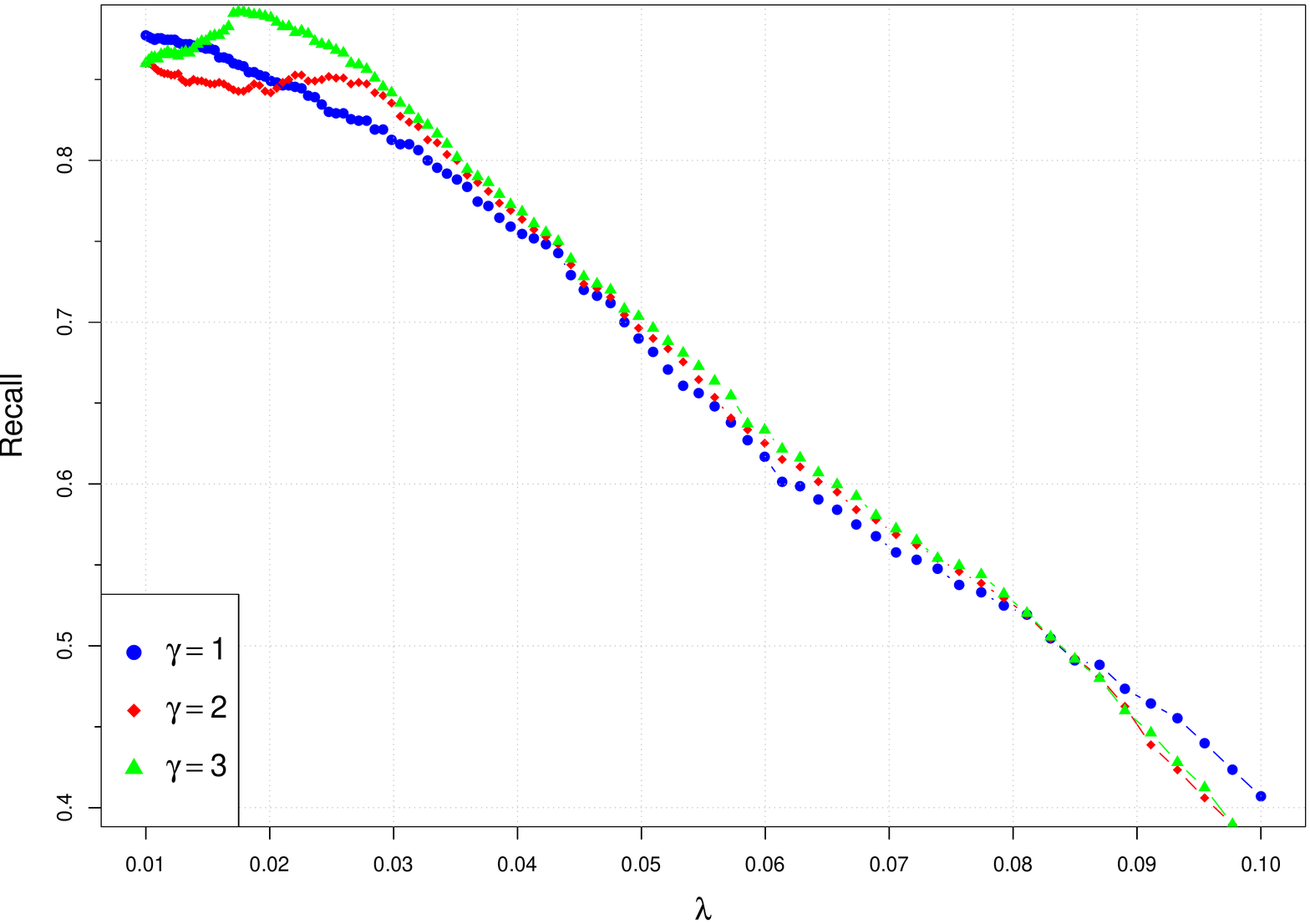}
		& \includegraphics[width=0.45\textwidth]{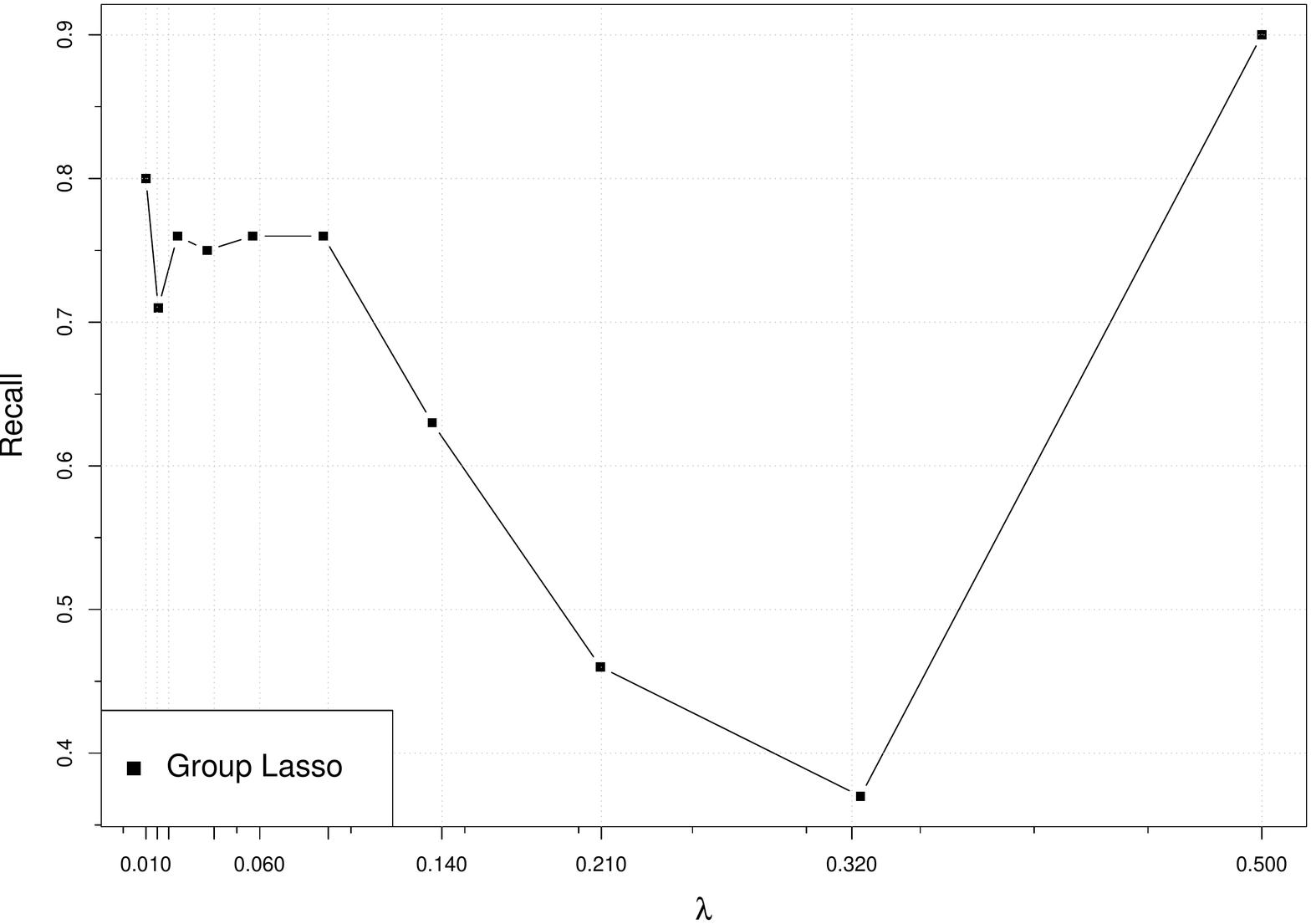} \\
		{\footnotesize (c) Recall of PISTA under different regularizations.}
		&  {\footnotesize (d) Recall of group Lasso under different regularizations.}
		\\
		\includegraphics[width=0.45\textwidth]{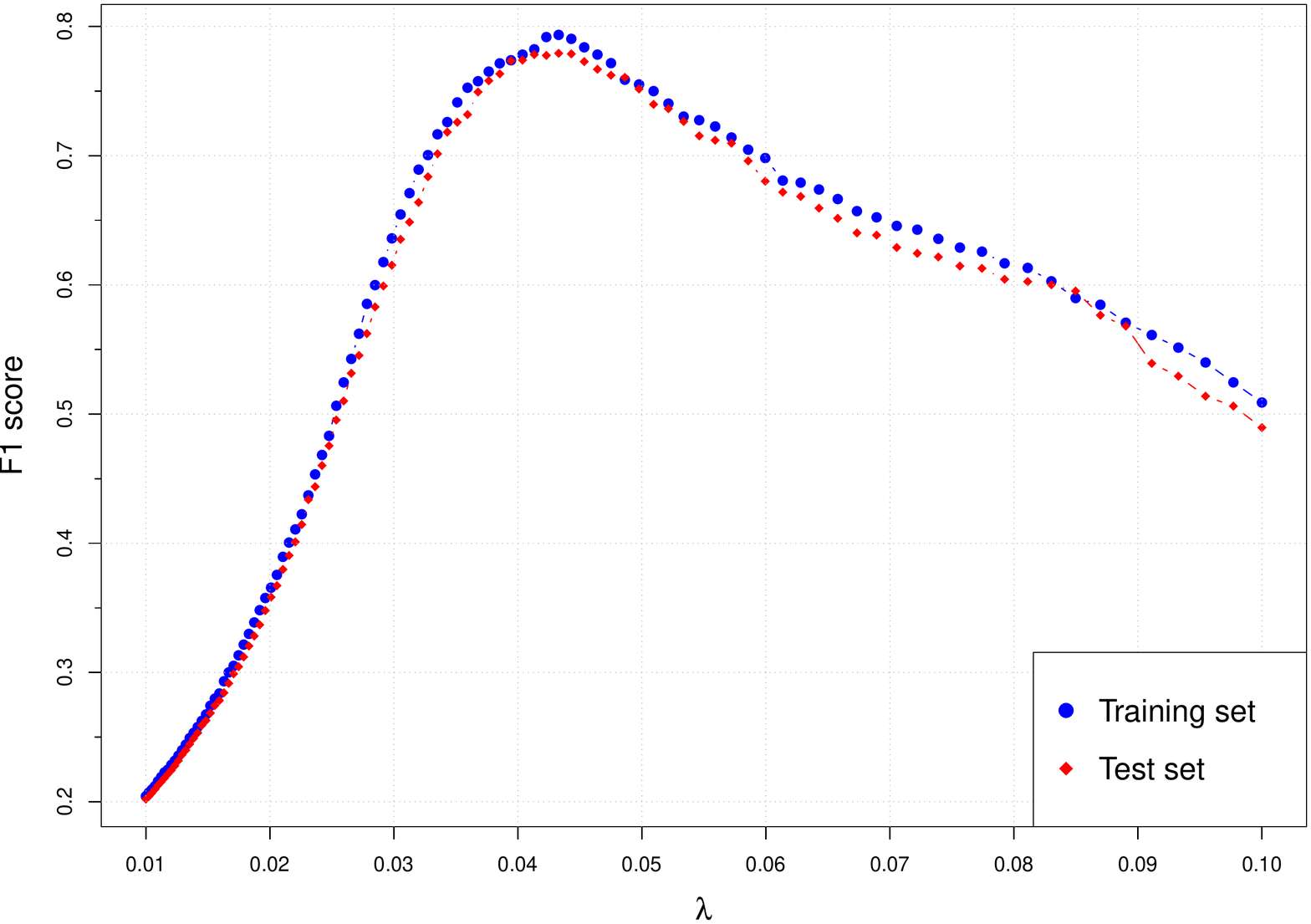}
		& \includegraphics[width=0.45\textwidth]{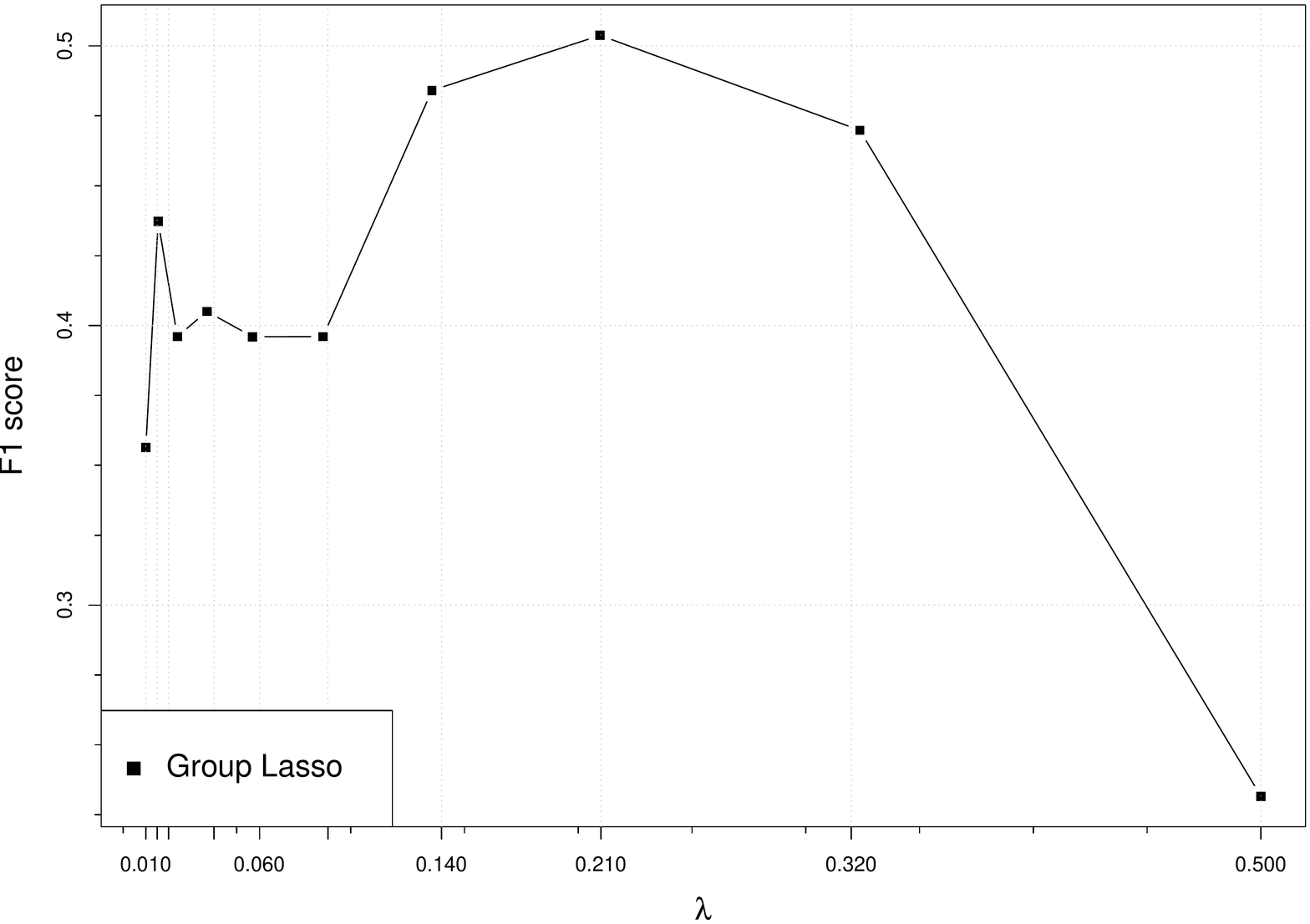} \\
		{\footnotesize (e) F1 score of PISTA under different regularizations.}
		&  {\footnotesize (f) F1 score of group Lasso under different regularizations.}
	\end{tabular}
	\caption{PISTA and group Lasso: a side-by-side comparison.}
	\label{fig::comparison}
\end{figure}

\begin{itemize}
	\item Figure \ref{fig::comparison} (a) and (c) show the performance comparison of PISTA with group MCP regularizer under different $\gamma$'s. It can be seen that, while smaller $\gamma$ makes the problem highly nonconvex, the precision and recall are not significantly affected by varying the coefficient.
	\item The optimal $\lambda$ that maximizes the F1 score for PISTA is between $0.04$ and $0.05$, and the generalization from the training set to test set can be seen in Figure \ref{fig::comparison} (e) for $\gamma = 1$.
	\item The precision and recall as well as the F1 score performance for group Lasso achieved using SAM package are shown in Figures \ref{fig::comparison} (b), (d), and (f). The optimal $\lambda$ that maximizes the F1 score for group Lasso is around $0.21$. Compared to PISTA, the convex counterpart has significantly less capability of maximizing the F1 score, which in return demonstrates the advantage of adopting nonconvex regularization.
\end{itemize}

We also list the complete result for the stock market data. This include the complete list of labels and their corresponding companies, as well as the set of causal relationships with sparsity being no more than 3. The results are shown in Tables \ref{tab::tickers_comp} and \ref{tab::stock}.

\begin{table}[!t]
	\centering
	\begin{tabular}{|c|c||c|c|}
		\hline 	Label & Company Name & Label & Company Name \\\hline
		ADBE & Adobe & AMD & Advanced Micro Devices\\\hline
		A & Agilent Technologies & ADI & Analog Devices \\\hline
		AKAM & Akamai Technologies & ALTR &  Altair Engineering\\\hline
		AMAT & Applied Materials & ADSK & Autodesk \\\hline
		ADP & Automatic Data Processing & AAPL & Apple \\\hline
		BMC & BMC Software & BRCM & Broadcom \\\hline
		CA & CA Technologies & CTXS & Citrix Systems \\\hline
		CSCO & Cisco Systems & CTSH & Cognizant \\\hline
		CSC & Computer Sciences Corporation & CPWR & Ocean Thermal Energy Corporation \\\hline
		DELL & Dell & EBAY & eBay \\\hline
		ERTS & Electronic Arts & EMC & Dell EMC \\\hline
		FFIV & F5 Networks & FIS & Fidelity National Information Services\\\hline
		FISV & Fiserv Inc & FLIR & FLIR Systems \\\hline
		HRS & Harris Corporation & HPQ & Hewlett Packard Enterprise \\\hline INTC & Intel & INTU & Intuit \\\hline 
		IBM & International Business Machines & JBL & Jabil Circuit \\\hline
		JDSU & JDS Uniphase Corporation & JNPR & Juniper Networks \\\hline
		KLAC & KLA-Tencor & LSI & LSI Corporation \\\hline
		LXK & Lexmark & LLTC & Linear Technology Corporation \\\hline
		WFR & MEMC Electronic Materials & MCHP & Microchip Technology \\\hline
		MOLX & Molex Inc & MSI & Motorola Solutions \\\hline
		MU & Micron Technology & MSFT & Microsoft Corporation \\\hline
		NSM & National Semiconductor & NTAP & NetApp \\\hline
		NVLS & Novellus Systems & NFLX & Netflix  \\\hline
		NVDA & Nvidia & ORCL & Oracle Corporation \\\hline PAYX & Paychex Inc & QCOM & Qualcomm \\\hline
		RHT & Red Hat Inc & SNDK & SanDisk \\\hline 
		SYMC & Symantec Corporation & TLAB & Tellabs Inc \\\hline TER & 	Teradyne Inc & TXN & Texas Instruments \\\hline TSS & Total System Services & VRSN & Verisign Inc \\\hline
		WDC & Western Digital & XRX & Xerox Corporation \\\hline XLNX & Xilinx Inc & YHOO & Yahoo Inc\\\hline 		
	\end{tabular}
	\caption{Complete set of labels and corresponding company names.}
	\label{tab::tickers_comp}
\end{table}
\begin{longtable}{|l|l|l|}
		\hline 	Company Label & Causal Relationship (PISTA) & Causal Relationship (glmnet) \\\hline
		ADBE & ALTR, ADSK, FLIR & FLIR, JDSU, KLAC \\\hline
		AMD & KLAC, MU, XRX & ERTS, FLIR, KLAC \\\hline
		A & FLIR, JDSU, NFLX & ERTS, FLIR, JDSU \\\hline
		AKAM & EBAY, NSM, ORCL, SYMC & A, EBAY, KLAC, SNDK \\\hline
		ALTR & DELL, NVLS, SNDK, WDC, XRX & ADBE, BRCM, FLIR, NVDA, WDC \\\hline
		ADI & MU, QCOM, SNDK & FLIR, QCOM, TSS \\\hline
		AAPL & AAPL, DELL, FLIR & AKAM, FLIR, TSS \\\hline
		AMAT & CA, DELL, FLIR & CPWR, FLIR, KLAC \\\hline
		ADSK & JNPR, ORCL, SNDK, TSS & ADSK, ADP, SNDK, TSS \\\hline
		ADP & ADI, INTU, SYMC, WDC & ADP, NFLX, PAYX \\\hline
		BMC & CA, CTSH, FLIR & ADSK, FIS, SNDK \\\hline
		BRCM & ADI, JDSU, LSI & JDSU, LSI, SNDK \\\hline
		CA & CA, DELL, FLIR, MU, MSI & ADSK, FIS, INTU, SNDK, TSS \\\hline
		CSCO & CA, FLIR, JDSU, NVDA & FLIR, JDSU, KLAC, NVDA \\\hline
		CTXS & AKAM, FISV, FLIR & AKAM, FLIR, WDC \\\hline
		CTSH & ADBE, AMD, MOLX, TSS, YHOO & FLIR, PAYX, RHT, WDC, YHOO \\\hline
		CSC & ADP, EMC, QCOM, YHOO & CPWR, QCOM, TSS, YHOO \\\hline
		CPWR & KLAC, SNDK, TSS & KLAC, NVDA, TSS \\\hline
		DELL & ADI, FLIR, INTU, QCOM & ADBE, QCOM, SNDK, TSS \\\hline
		EBAY & ADBE, FFIV, JNPR, MU, YHOO & AMAT, JNPR, KLAC, SYMC \\\hline
		ERTS & MU, TSS, WDC & HRS, TSS, WDC \\\hline
		EMC & CA, CTSH, HPQ & ADSK, KLAC, SNDK \\\hline
		FFIV & BRCM, CA, MU, YHOO & BRCM, MSI, NVDA, YHOO \\\hline
		FIS & ADBE, AAPL, TLAB & HRS, KLAC, RHT \\\hline
		FISV & AKAM, DELL, FLIR, NVLS & AKAM, FLIR, MU, TSS \\\hline
		FLIR & ADBE, ADSK, TSS & JDSU, MU, WDC \\\hline
		HRS & ADSK, CTXS, CSC, EMC & INTU, NVLS, TSS, WDC\\\hline
		HRS & A, ADSK, DELL, FLIR, JBL & INTU, NVLS, TSS, WDC \\\hline
		HPQ & A, ADSK, DELL, FLIR, JBL & FIS, FLIR, INTU, KLAC, TSS \\\hline
		\multirow{2}{*}{}INTC & AKAM, DELL, FFIV, &  FLIR, JDSU, MU, \\ & FLIR, INTU, KLAC & WDC \\\hline
		INTU & FLIR, JBL, NVDA, PAYX & FLIR, JBL, MSI \\\hline
		JBL & FFIV, FLIR, XRX & FLIR, KLAC, WDC \\\hline
		JDSU & DELL, FLIR, PAYX & FLIR, KLAC, TSS \\\hline
		JNPR & ADBE, CTSH, FLIR, MU, NVDA, ORCL & ADBE, KLAC, SNDK, TER \\\hline
		KLAC & ADBE, DELL, FLIR, INTU & FLIR, QCOM, TSS \\\hline
		LXK & ADBE, CA, NVLS, NVDA, SNDK & KLAC, NVDA, TSS \\\hline
		\multirow{2}{*}{LLTC} & AKAM, DELL, INTU, & FLIR, JDSU, SNDK,\\ & QCOM, SNDK, TLAB & TSS, WDC \\\hline
		LSI & CA, NFLX, TSS & CA, INTU, MSI \\\hline
		WFR & ADBE, CA, RHT & ADBE, KLAC, QCOM \\\hline
		MCHP & ADBE, AKAM, FFIV, JDSU & FLIR, JDSU, QCOM \\\hline
		MU & AKAM, EMC, MU & RHT, SNDK, XRX \\\hline
		MSFT & DELL, NTAP, SNDK & DELL, FLIR, NVDA \\\hline
		MOLX & DELL, FLIR, MU, TSS & FLIR, JDSU, MU, TSS\\\hline
		\multirow{2}{*}{MSI} & AMD, AKAM, ADP, &  AKAM, FIS, JDSU, \\ & BRCM, EMC, INTU & VRSN, YHOO \\\hline
		NSM & JDSU, NVDA, QCOM & FLIR, JDSU, QCOM \\\hline
		NTAP & ADP, JDSU, MU, SNDK & ADP, FLIR, JDSU, SNDK \\\hline 
		\multirow{2}{*}{NFLX} & BMC, BRCM, EBAY, EMC, FIS, & FFIV, FLIR, HPQ, \\ & FLIR, HPQ, MSI, NVLS & ORCL, SNDK, TSS \\\hline
		NVLS & ADBE, FFIV, JDSU & FLIR, JDSU, QCOM \\\hline
		NVDA & ADI, MU, QCOM & LLTC, QCOM, WDC \\\hline
		\multirow{2}{*}{ORCL} & CA, FFIV, FLIR, & AAPL, CA, JDSU, \\ & JDSU, NVDA, WDC & MSI, SNDK, WDC \\\hline
		PAYX & JBL, JDSU, MCHP, MU & EBAY, JDSU, MU \\\hline
		QCOM & A, FLIR, KLAC, WDC & AAPL, FLIR, TLAB, WDC \\\hline
		RHT & ADSK, DELL, INTC & ADSK, ADP, NVDA \\\hline
		SNDK & ADI, CA, FLIR, MU & BRCM, KLAC, NVDA \\\hline
		SYMC & FFIV, INTU, NFLX & CSC, EMC, WDC \\\hline
		TLAB & CA, FLIR, JDSU, NVDA & FLIR, JDSU, NVDA \\\hline
		\multirow{2}{*}{TER} & BRCM, CTXS, FLIR, & FLIR, JDSU, KLAC, \\ & JDSU, JNPR, KLAC &  NVDA \\\hline
		TXN & CA, DELL, FLIR, QCOM & ADSK, FIS, FLIR, TSS \\\hline
		\multirow{2}{*}{TSS} & CA, FLIR, HPQ, & FLIR, HRS, INTU, \\ & JBL, JDSU, PAYX, YHOO & JBL, KLAC, TSS \\\hline
		VRSN & LSI, MCHP, MU & AAPL, LSI, TLAB \\\hline
		WDC & LSI, NVDA, VRSN & AMD, NVDA, YHOO \\\hline
		XRX & ADI, TLAB, TSS, XRX & EBAY, TLAB, TSS, XRX \\\hline
		XLXN & DELL, FLIR, MU & FLIR, SNDK, WDC \\\hline
		YHOO & ADBE, FLIR, JDSU, YHOO & ADBE, JDSU, KLAC, TSS\\\hline
	\caption{Complete set of causal relationships recovered by PISTA and glmnet with sparsity limited to 3. For each row, the companies in the second (or third) column causally affect the company in the first column. For some rows, the number of companies that causal link is strong in the sense that the impact is nonzero even for large values of $\lambda$. For these companies, we keep the causal relations acquired under the largest $\lambda$, which is $0.1$ for PISTA.}
	\label{tab::stock}
\end{longtable}

\end{document}